\pdfoutput=1

\documentclass[11pt]{article}

\usepackage{ACL2023}

\usepackage{times}
\usepackage{latexsym}

\usepackage[T1]{fontenc}

\usepackage[utf8]{inputenc}

\usepackage{microtype}

\usepackage{inconsolata}

\usepackage{hyperref}
\usepackage{amsmath}
\usepackage{amssymb}
\usepackage{mathtools}
\usepackage{amsthm}
\usepackage[capitalize,noabbrev]{cleveref}

\usepackage[textsize=tiny]{todonotes}

\newcommand*\samethanks[1][\value{footnote}]{\footnotemark[#1]}

\usepackage{times}
\usepackage{latexsym}
\usepackage[T1]{fontenc}
\usepackage[utf8]{inputenc}
\usepackage{inconsolata}
\usepackage{array,colortbl,multirow,multicol,booktabs,ctable, tablefootnote, lipsum}
\newcommand{\minisection}[1]{\noindent{\bf #1}\hspace{0.6em}}
\usepackage{pgfplots}
\usepgfplotslibrary{colormaps}
\usetikzlibrary{pgfplots.groupplots}
\pgfplotsset{compat=1.3}
\usepackage{tikz}
\usetikzlibrary{patterns}
\usepgfplotslibrary{colorbrewer}
\usepgfplotslibrary{groupplots}

\title{Faithful Chain-of-Thought Reasoning}

\author{
  Qing Lyu \thanks{\hspace{4pt} Equal contribution.} \hspace{5mm}
  Shreya Havaldar\samethanks \hspace{5mm}
  Adam Stein\samethanks \hspace{5mm}
  Li Zhang\\
  \bf{Delip Rao} \hspace{5mm}
  \bf{Eric Wong} \hspace{5mm}
  \bf{Marianna Apidianaki} \hspace{5mm}
  \bf{Chris Callison-Burch} \\
  University of Pennsylvania \\
  \texttt{\{lyuqing, shreyah, steinad, zharry, exwong,}\\
  \texttt{marapi, ccb\}@seas.upenn.edu, deliprao@gmail.com}
}

\begin{document}
\maketitle
\begin{abstract}
While Chain-of-Thought (CoT) prompting boosts Language Models' (LM) performance on a gamut of complex reasoning tasks, the generated reasoning chain does not necessarily reflect how the model arrives at the answer (aka. \emph{faithfulness}). We propose \textbf{Faithful CoT}, a reasoning framework involving two stages: \textbf{Translation} (Natural Language query $\rightarrow$ symbolic reasoning chain) and \textbf{Problem Solving} (reasoning chain $\rightarrow$ answer), using an LM and a deterministic solver respectively. This guarantees that the reasoning chain provides a faithful explanation of the final answer. Aside from interpretability, Faithful CoT also improves empirical performance: it outperforms standard CoT on 9 of 10 benchmarks from 4 diverse domains, with a relative accuracy gain of 6.3\% on Math Word Problems (MWP), 3.4\% on Planning, 5.5\% on Multi-hop Question Answering (QA), and 21.4\% on Relational Inference. Furthermore, with GPT-4 and Codex, it sets the new state-of-the-art few-shot performance on 7 datasets (with 95.0+ accuracy on 6 of them), showing a strong synergy between faithfulness and accuracy.\footnote{Our code, data, and prompts are available at \url{https://github.com/veronica320/Faithful-COT}.}
\end{abstract}

\section{Introduction}
\label{sec:intro}

Complex reasoning tasks, such as commonsense reasoning and math reasoning, have long been the Achilles heel of LMs \cite{bengio_system_2019}, until a recent line of work on Chain-of-Thought (CoT) reasoning (\citealp[i.a.]{wei_chain_2022, wang_self-consistency_2022}) brought striking performance gains. CoT prompts an LM to generate a reasoning chain along with the answer, given only a few in-context exemplars. 

\begin{figure}[t]
\centering
\includegraphics[width=0.95\columnwidth]{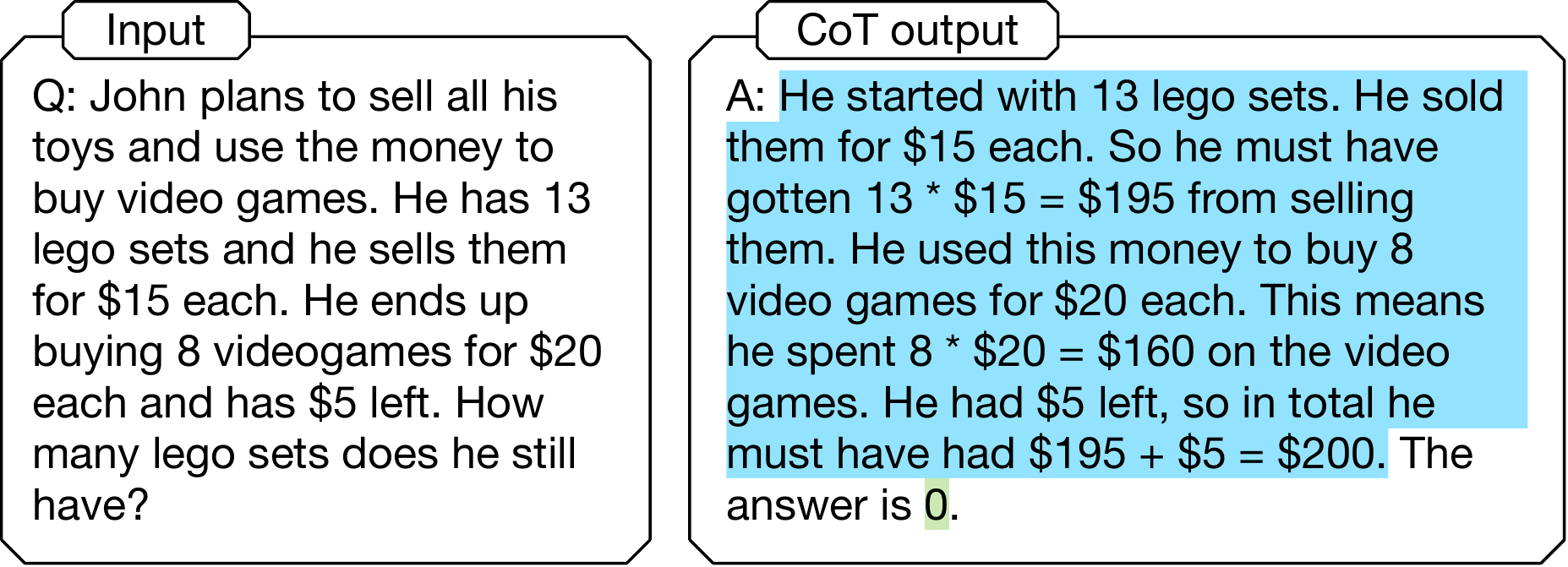}
\vspace{-0.05in}
\caption{An example of \textit{unfaithful} output from CoT prompting \cite{wei_chain_2022} on GSM8K. The answer (green) does not follow from the reasoning chain (blue).}
\vspace{-0.1in}
\label{fig:cot_unfaithful}
\end{figure}

Besides performance improvement, CoT is also claimed to ``provide an interpretable window into the behavior of the model'' \cite{wei_chain_2022}. However, it is unclear how accurately these reasoning chains reflect the underlying reasoning process behind the model's prediction, namely, how \emph{faithful} they are as explanations \cite[i.a.]{jacovi_towards_2020}. 
In standard CoT, faithfulness is not guaranteed and even systematically violated \cite{turpin_language_2023}, as the final answer does not necessarily follow from the generated reasoning chain.
In other words, CoT can ``lie'' about the model's true reasoning process.
Figure~\ref{fig:cot_unfaithful} exemplifies such an unfaithful CoT generation from \citet{wei_chain_2022} on GSM8K:
the answer ``0'' is not even mentioned in the reasoning chain. This, along with more examples in  Appendix~\ref{appendix:unfaithful_output}, illustrates that standard CoT does not provide interpretability of how the model predicts the answer.

The lack of faithfulness in CoT can be dangerous in high-stake applications because it may mislead people into believing that the model is self-interpretable, while there is no actual causal relationship between the reasoning chain and the answer. Even worse, when an \textit{unfaithful} explanation looks \textit{plausible} (i.e., convincing to humans) \cite{jacovi_towards_2020}, this makes it easier for people (e.g., legal practitioners)  to over-trust the model (e.g., a recidivism predictor) even if it has implicit biases (e.g., against racial minorities) \cite{pruthi_learning_2020, slack_fooling_2020}. 

\begin{figure}
\centering
\includegraphics[width=0.5\columnwidth]{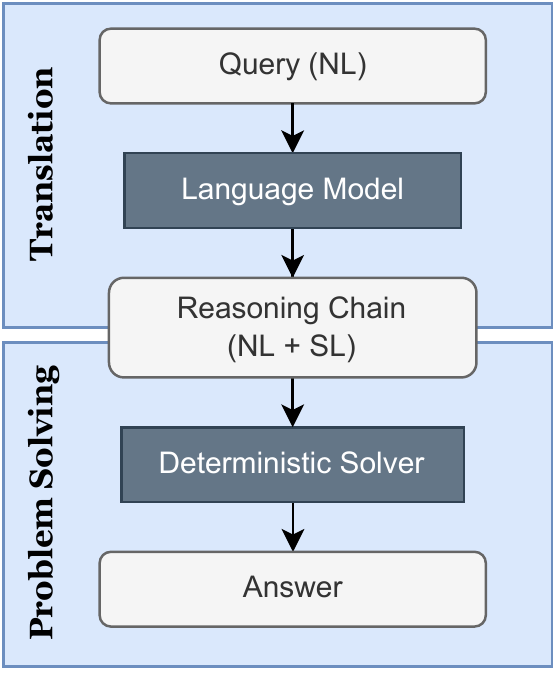}
\caption{An overview of our Faithful CoT framework, consisting of \textbf{Translation}, where an LM translates a query (in NL/Natural Language) into a reasoning chain (which interleaves NL and SL/Symbolic Language), and \textbf{Problem Solving}, where an external solver executes the reasoning chain to derive the answer. }
\vspace{-0.2in}
\label{fig:2Stage}
\end{figure}

To address this concern, we propose \textbf{Faithful CoT}, a reasoning framework where the answer is the result of deterministically executing the reasoning chain. Specifically, we break down a complex reasoning task into two stages: \textbf{Translation} and \textbf{Problem Solving} (Figure~\ref{fig:2Stage}). During Translation, an LM translates a query into a reasoning chain, which interleaves NL and Symbolic Language (SL). The NL component decomposes the original query into multiple simpler, interdependent subproblems. Then, each subproblem is tackled in a task-dependent SL, such as Python, Datalog, or Planning Domain Definition Language (PDDL). In the Problem Solving stage, the reasoning chain is executed by a deterministic solver, e.g., a Python/Datalog interpreter, or a PDDL planner, to derive the answer.

Our reasoning chain (outcome of Translation) is guaranteed to provide a faithful explanation of how the final answer is produced (outcome of Problem Solving), therefore making our method more \textit{interpretable} than standard CoT methods.\footnote{Note that we do not claim that the process of generating the reasoning chain itself, i.e., the Translation stage, is interpretable. See more discussion in Limitations.} While \textbf{interpretability is not the same as correctness}  (i.e. our method can reveal the reasoning process behind \textit{both} correct \textit{and} wrong answers), we find that \textbf{it does empirically improve correctness}: when evaluated on 10 reasoning datasets from 4 diverse domains (MWP, Planning, Multi-hop QA, and Relational Inference), Faithful CoT brings consistent performance gains over three existing baselines, across different LMs and decoding strategies. With Codex, our approach outperforms vanilla CoT on 9 of the 10 datasets, with a relative accuracy gain of 6.3\% on MWP, 3.4\% on Planning, 5.5\% on Multi-hop QA, and 21.4\% on Relational Inference. With GPT-4, our method sets the new SOTA few-shot performance on 7 datasets, with 95.0+ accuracy on 6 of them. This suggests that interpretability does not have to come at the cost of performance; instead, there exists a strong synergy in between.

Our key contributions are as follows:\\
(a) We propose Faithful CoT, a framework that decomposes reasoning into Translation and Problem Solving. The reasoning chain interleaves user-understandable natural language comments and executable symbolic language programs, thus providing faithful interpretability of how the model arrives at the answer.\\
(b) Our approach is generalizable to multiple domains beyond arithmetic reasoning and simple symbolic reasoning, thanks to its flexible integration with any choice of SL and external solver. We set the new SOTA performance on 7 out of the 10 reasoning datasets, showing a strong synergy between faithfulness and accuracy.\\
(c) We provide an extensive analysis of the strengths and weaknesses of our method, showing its generalizability across LMs, robustness to the choice of exemplars and prompt phrasing, the pivotal role of the solver, the plausibility of generated reasoning chains, as well as frequent error patterns where it still struggles.

\section{Related Work}
\label{sec:related_work}

\minisection{Faithfulness.} In interpretability, \textit{faithfulness} (also called \textit{fidelity} or \textit{reliability}) means that an explanation should ``accurately represent the reasoning process behind the model's prediction'', which is a fundamental requirement of an explanation \cite{harrington_harvey_1985, ribeiro_why_2016, gilpin_explaining_2018, jacovi_towards_2020}.\footnote{Note that this differs from the notion of faithfulness in the Natural Language Generation (NLG) literature, primarily in what constitutes the \textbf{ground truth}. In interpretability, we talk about the faithfulness of an explanation w.r.t. \textbf{the model's underlying reasoning mechanism} -- the ground truth is usually unknown. In NLG, we talk about the faithfulness of the generated text (e.g., a translated sentence, or a summary) w.r.t. some \textbf{explicit source} (e.g., the source sentence, or the full document) -- the ground truth is transparent.} It should be contrasted with \textit{plausibility} (a.k.a. \textit{persuasiveness} or \textit{understandability}), which refers to ``how convincing an explanation is to humans'' \cite{herman_promise_2019, jacovi_towards_2020}. In the context of CoT prompting, a faithful reasoning chain needs to accurately reflect how the model arrives at the final answer, whereas a plausible reasoning chain is one that looks reasonable and coherent to humans. Standard CoT \cite{wei_chain_2022} generates the reasoning chain in pure NL, which may often look plausible; nevertheless, the final answer does not need to causally follow from the reasoning chain, thus not guaranteeing faithfulness.

\minisection{Chain-of-Thought-style prompting.} In CoT-style prompting, given a complex question $Q$, an LM is prompted to generate a reasoning chain $C$ along with the final answer $A$. Specifically, the prompt consists of a few examples of $(Q, C, A)$ triples, called in-context exemplars. This allows pre-trained LMs (e.g., GPT-3 \cite{brown_language_2020}) to solve unseen questions with much higher accuracy than standard prompting, where the exemplars do not contain the reasoning chain $C$. 

\begin{figure*}[t!]
\centering
\includegraphics[width=0.89\textwidth]{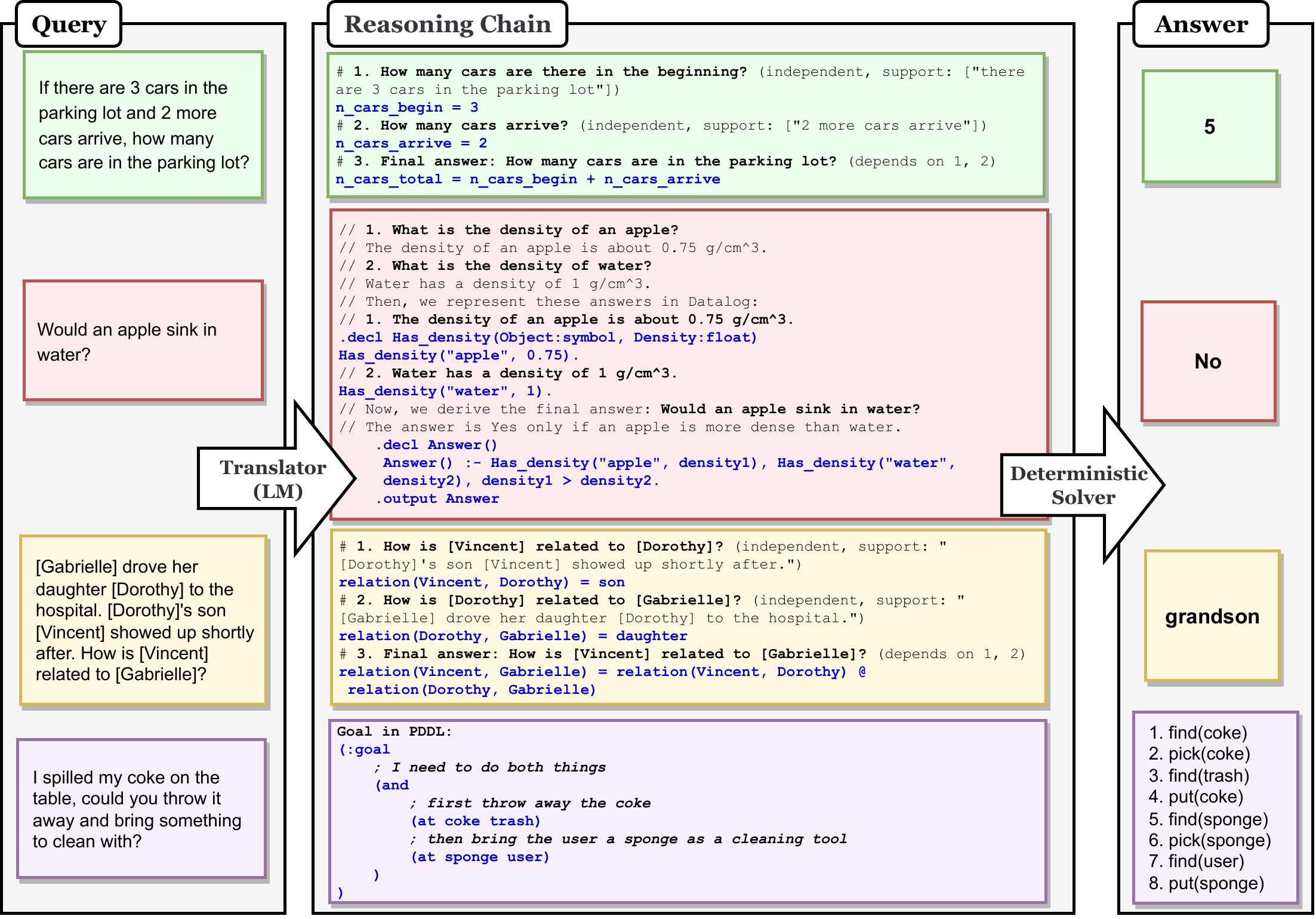}
\caption{Examples from each task (Math Word Problems, Multi-hop QA, Relational Inference, and Planning) showing our 2-stage Translation and Problem Solving pipeline.}
\vspace{-0.2in}
\label{fig:method}
\end{figure*}

We create a taxonomy of existing CoT-style prompting methods into three types: all-at-once, ensemble-based, and modularized. \textbf{All-at-once} prompting means that the LM produces $C$ and $A$ as one continuous string, without any dependencies or constraints in between. Scratchpad \cite{nye_show_2021}, standard CoT \cite{wei_chain_2022}, and ``Let's think step by step'' \cite{kojima_large_2022}, are all examples of this kind. \textbf{Ensemble-based} prompting is designed to overcome the local optimality issue of the one-shot generation in previous methods by sampling multiple $(C,A)$ pairs and choosing the best answer via strategies like majority voting. Examples include Self-Consistent CoT \cite{wang_self-consistency_2022}, Minerva \cite{lewkowycz_solving_2022}, and DIVERSE \cite{li_advance_2022}, which differ mainly in the voting granularity and the underlying LM. \textbf{Modularized} methods break down $Q$ into subproblems and then conquer them individually (\citealp[i.a.]{jung_maieutic_2022, qian_limitations_2022}). In particular, Least-to-Most prompting \cite{zhou_least--most_2022} has a similar question decomposition process to ours, but there is still no faithfulness guarantee since the reasoning chain is entirely in NL.

Concurrent with our work, \citet{chen_program_2022} and \citet{gao_pal_2022} both generate Python programs (i.e., SL-only reasoning chains) to derive the answer. We want to highlight the following qualitative differences:\footnote{Also see Appendix~\ref{sec:appendix_concurrent_work} for an empirical comparison.} (a) In terms of motivation, our approach is interpretability-driven, whereas theirs are performance-driven. (b) Our reasoning chain involves a structured decomposition of the problem in NL, allowing users without a programming background to better understand and potentially interact with the system. (c) They only use Python as the SL and only tackle math and simple symbolic reasoning tasks, whereas we demonstrate the generalizability of our approach to multiple symbolic languages and various other domains. In particular, we innovatively recast a diverse set of realistic tasks (Planning, Multi-hop QA, and Relational Inference) into a symbolic representation, which allows us to tackle them with a single framework. (d) We perform a more comprehensive analysis compared to previous work, especially a human evaluation of the reasoning chain correctness.

\section{Method}

Our method, \textbf{Faithful CoT}, is a 2-stage pipeline, as seen in Figure~\ref{fig:2Stage}. Like previous CoT-style work, our prompt consists of $(Q, C, A)$ triples. Notable differences lie in our unique interleaving of NL (natural language) and SL (symbolic language) in $C$, as well as the way we derive the final answer $A$.

In the \textbf{Translation} stage, given a complex query $Q$ in NL, we prompt an LM to translate it into a reasoning chain $C$, which interleaves NL comments and SL programs. The NL component decomposes the original query into multiple simpler, interdependent subproblems. Then, each subproblem is tackled in a task-dependent SL, such as Python, Datalog, or PDDL. In the \textbf{Problem Solving} stage, we call a deterministic external solver, e.g., a Python interpreter, a Datalog executor, or PDDL planner, depending on the task, to obtain the answer $A$ from the reasoning chain $C$. As shown in Figure~\ref{fig:method}, we define $C_{NL}$ to be the NL component (black) and $C_{SL}$ to be the SL component (blue) in $C$. Though we separate the two components notationally, they are interleaved in the generation. Using this approach, $C$ is guaranteed to be a faithful model explanation, since our final $A$ is the result of deterministically executing $C_{SL}$. Moreover, $C_{NL}$ allows the user to better understand the reasoning process.\footnote{While no constraints are enforced between $C_{NL}$ and $C_{SL}$ in our main experiments, we analyze this in Section~\ref{sec:appendix_constraint}.}

We apply this method to 4 types of complex reasoning tasks: MWP, Multi-hop QA, Planning, and Relational Inference. Next, we will illustrate how our method works for each of them, with examples from Figure~\ref{fig:method}. 

\subsection{Math Word Problems (MWP)} Given a grade-school math question $Q$ written in NL (``If there are 3 cars in the parking lot and 2 more cars arrive, how many cars are in the parking lot?'', shown in green in Figure~\ref{fig:method}), we want to obtain $A$ as a real-valued number (\texttt{5}). In the Translation stage, we prompt the LM to take in $Q$ and generate a reasoning chain $C$, which interleaves $C_{NL}$ and $C_{SL}$. Specifically, the $C_{NL}$ component consists of three types of information: 

(a) \textbf{Subquestions}: $Q$ is broken down into multiple smaller-scale subquestions, e.g., ``1. how many cars are there in the beginning?'', ``2. how many cars arrive?'', and ``3. how many cars are in the parking lot?''.

(b) \textbf{Dependency Graph}: Each subquestion can either be answered directly via context (subquestions 1 and 2 are ``independent'') or rely on answers to previous subquestions (subquestion 3 ``depends on 1 and 2'').

(c) \textbf{Rationales}: Each subquestion is accompanied with 
 rationale(s) to support the answer (the ``support'' field). The rationales can be either a subset of the original context (``2 more cars
arrive'') or any external knowledge (``there are 7 days in a week'') relevant to the subquestion.

Each subquestion and its corresponding dependencies and rationales inform the subsequent generation of  $C_{SL}$. In our example in Figure~\ref{fig:method}, $C_{SL}$ consists of Python code generated to answer each subquestion in $C_{NL}$. During the Problem Solving stage, we execute $C_{SL}$ using our solver, a Python interpreter, to derive $A$ (5 cars in the end).

\subsection{Multi-hop QA} Given a complex question $Q$ that involves multiple steps of reasoning (e.g., ``Would a pear sink in water?'', shown in red in Figure~\ref{fig:method}), we want to obtain the answer $A$ as a Boolean value or string value variable. Similar to our MWP task formulation, $C$ interleaves $C_{NL}$ (NL comments), and $C_{SL}$ (symbolic program). Depending on the nature of the task, the format of the reasoning chain $C$ is slightly different: for some datasets, the LM first generates all subquestions and their answers in NL, and then represents these answers as SL to derive $A$ (see Figure~\ref{fig:method}); for others, the LM interleaves the NL subquestions and the SL program, similar to the case of MWP (see Table~\ref{tab:appendix-date_prompt} and Table~\ref{tab:appendix-sports_prompt} for examples).
In terms of SL, we use both Python and Datalog, also depending on the dataset. As Multi-hop QA problems involve multi-step reasoning to solve, $C_{SL}$ often utilizes Boolean algebra and string comparisons (in Python) along with relation definitions and logic programming (in Datalog). We use their corresponding interpreter as our deterministic solver to execute $C_{SL}$ and obtain $A$. 

In the example from Figure~\ref{fig:method}, the LM first generates the subquestions, ``1. What is the density of a pear?'' and ``2. What is the density of water?'', which are individually answered in NL. The answers (``Water has a density of $1 g/cm^3$'') are converted to Datalog statements (\texttt{Has\_density(``water'', 1)}), which are then combined to formalize the truth condition of the final answer. Finally, we execute the Datalog program to determine that a pear would \textbf{not} sink in water.

\subsection{Planning} 

In a user-robot interaction scenario, given a household task query $Q$ from a user, we want to come up with a plan of actions $A$ that the robot should take in order to accomplish the task. For example, in Figure~\ref{fig:method}, given user query ``I spilled my coke on the table, could you throw it away and bring something to clean with?'', a possible plan can be ``find(coke), pick(coke), find(trash), put(coke) ...''. In the Translation stage, an LM translates $Q$ into $C$, consisting of $C_{NL}$ (which breaks down $Q$ into subtasks) and $C_{SL}$ (which represents the subtasks as a symbolic \texttt{goal} in PDDL\footnote{\url{https://en.wikipedia.org/wiki/Planning_Domain_Definition_Language}. A \texttt{goal} is a special construct in PDDL.} --- a language to define and solve classical planning problems).
Figure~\ref{fig:method} shows this translation, with $C_{SL}$ in blue and $C_{NL}$ in black. Finally, we call a PDDL Planner as the deterministic solver to obtain $A$, a plan to accomplish the goal $C_{SL}$ under the predefined scenario.


\subsection{Relational Inference}
\label{sec:relational_inference}
Given a relational inference problem $Q$ written in NL, we want to obtain $A$ as a string-valued variable. For example, the \textbf{CLUTRR} \cite{sinha_clutrr_2019} dataset involves inferring the family relationship (e.g., ``grandson'') between two people from a short story (e.g., ``[Gabrielle] drove her daughter [Dorothy] to the hospital. [Dorothy]'s son [Vincent] showed up shortly after. How is [Vincent] related to [Gabrielle]?'', shown in yellow in Figure~\ref{fig:method}). During the Translation stage, we prompt the LM to generate $C$, consisting of $C_{NL}$ and $C_{SL}$. Similar to previous tasks, $C_{NL}$ breaks down $Q$ into subquestions (``How is [Vincent] related to [Dorothy]'' and ``How is [Dorothy] related to [Gabrielle]''), as well as provide input extracts as rationales to support the answer (``[Dorothy]'s son [Vincent] showed up shortly after'', etc.). Each subquestion in $C_{NL}$ is answered in $C_{SL}$ via a relational expression representing the relation between the mentioned entities, for example, \texttt{relation(Vincent, Dorothy)=son} denotes that Vincent is Dorothy's son. 
In the Problem Solving stage, our solver is a simple relational inference engine that relies on a set of transitivity rules provided by \citet{zhang2022improved} among possible family relationships, e.g., \texttt{son@daughter=grandson} (the son of one's daughter is one's grandson). Our solver recursively applies these rules on $C_{SL}$ to derive $A$, and determine that Vincent is Gabrielle's grandson.

\section{Experimental setup}

\begin{figure*}[t!]
\centering
\includegraphics[width=0.95\textwidth]{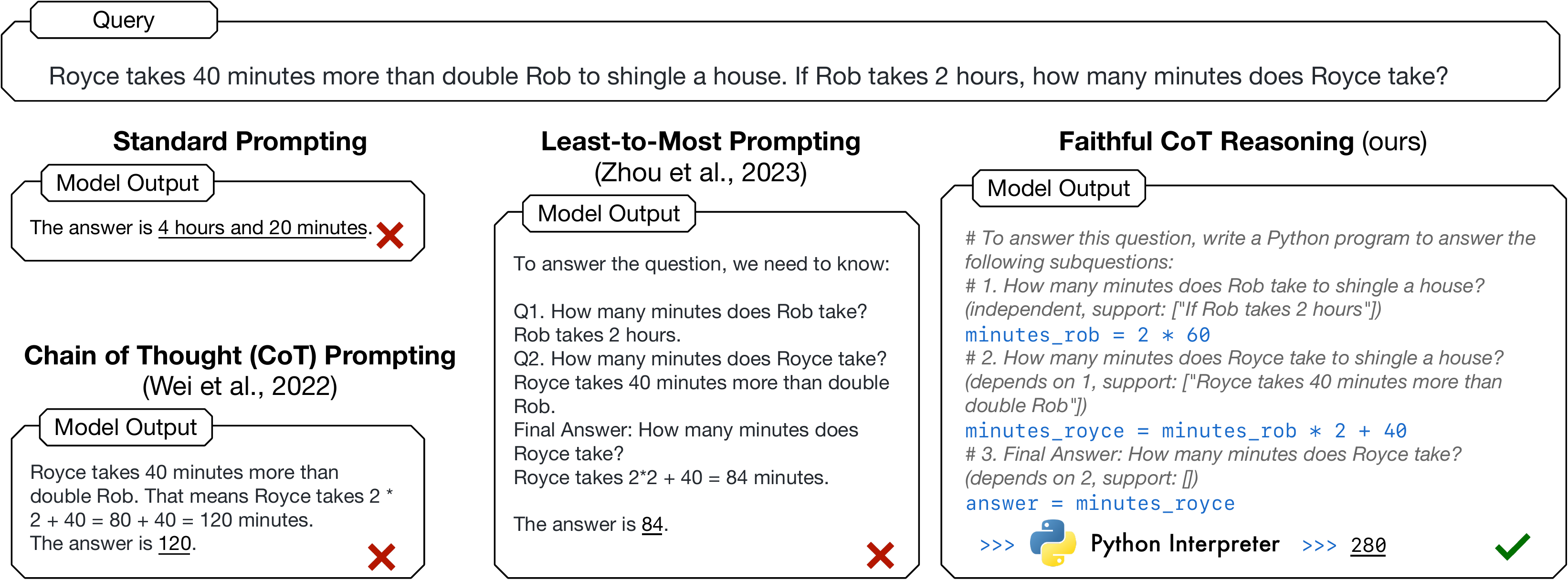}
\caption{A sample output for a math question from three baselines and Faithful CoT (our method). The ground-truth answer is 280, and only our method correctly computes the answer.}
\vspace{-0.13in}
\label{fig:baseline_and_our_prompts}
\end{figure*}

\subsection{Datasets}\label{sec:datasets} Here, we summarize the evaluation datasets used for each domain. We select the same number (6 to 10, depending on the task) of exemplars as in \citet{wei_chain_2022} to form our few-shot prompt, which can be found in our repository. Unless otherwise stated, we use the official splits: training set for exemplar selection, validation set for prompt tuning, and test set for evaluation.\footnote{See Appendix~\ref{sec:appendix_dataset_details} for dataset statistics, examples, data cleaning method, splits, prompt construction strategy, etc.}

\minisection{Math Word Problems (MWP).} We follow \citet{wei_chain_2022} and consider the same five MWP benchmarks: \textbf{GSM8K} \cite{cobbe_training_2021}, \textbf{SVAMP} \cite{patel_are_2021}, \textbf{MultiArith} \cite{roy_solving_2015}, \textbf{ASDiv} \cite{miao_diverse_2020}, and \textbf{AQuA} \cite{ling_program_2017}. For all datasets, the input question is phrased in NL. The answer is a string-valued mathematical expression for AQuA, and one or more integer(s) for all other datasets. We use the same 8-shot prompt for all datasets except AQuA.

\minisection{Multi-hop QA.} We consider the three datasets: \textbf{StrategyQA} \cite{geva_did_2021}, a dataset of open-domain questions that require an implicit multi-step strategy to answer, e.g., ``Did Aristotle use a laptop?'' involves answering ``1. When did Aristotle live?'', ``2. When was the laptop invented?'', and ``3. Is \#2 before \#1?''; \textbf{Date Understanding} from BIG-bench \cite{big-bench_collaboration_beyond_2021}, which asks the model to infer a date from a context, by performing computation on relative periods of time; and finally, \textbf{Sports Understanding} from BIG-bench, which asks the model to decide whether an artificially constructed statement related to sports is plausible or implausible. Since the latter two datasets do not have a training set, we follow \citet{wei_chain_2022} and select 10 examples from the test set to form the prompt and use the rest for evaluation.

\minisection{Planning.} We use the \textbf{SayCan} dataset \cite{ahn_as_2022}, which assumes a scenario of a robot operating in a kitchen, helping the user with household tasks, e.g., ``bring a coke to the table''. There are a number of locations and objects that the robot can interact with. The robot can only perform a fixed set of actions, including \texttt{find}, \texttt{pick}, and \texttt{put}. The task is to map a user query in NL to a plan of predefined actions. Following \citet{wei_chain_2022}, we manually write 7 exemplars, since no training set is provided.

\minisection{Relational inference.} We use the \textbf{CLUTRR} \cite{sinha_clutrr_2019} benchmark described in Section~\ref{sec:relational_inference}. The dataset has multiple splits based on the number of intermediate steps $K$ required to reach the answer. We construct the prompt using 8 exemplars with $K \in \{2,3\}$, and test the models on the remaining examples with $K$ up to 10.

\subsection{Evaluation Metrics} We evaluate the model performance with final answer accuracy as the main metric. Following previous work \cite{wei_chain_2022, wang_self-consistency_2022, chen_program_2022}, for all MWP datasets (except AQuA) where the answer contains integer(s), a correct answer is defined as the exact match between the prediction and the ground truth both converted to the nearest integer; for StrategyQA and Sports Understanding where the answer is a Boolean value, it is defined as the exact match between the prediction and the ground truth both evaluated as a Boolean variable; for SayCan, the generated plan is considered correct if it is among the ground truth plans; for all other datasets, we rely on the exact match between the prediction string and the ground truth string. Additionally, we evaluate the human-rated plausibility of the reasoning chain in Appendix~\ref{sec:appendix_human_eval_details}.

\subsection{Baselines} We compare our method to three other few-shot prompting baselines, shown in Figure~\ref{fig:baseline_and_our_prompts}: \textbf{standard} prompting, popularized by \citet{brown_language_2020}, with demonstrations of only the question and the answer;  \textbf{CoT} \cite{wei_chain_2022}, which additionally includes an NL reasoning chain; and \textbf{Least-to-Most (LtM)} \cite{zhou_least--most_2022}, which decomposes the question in NL but does not involve SL. 
All prompting methods are compared under two decoding strategies: \textbf{greedy} decoding, where the LM samples the most probable next token from the vocabulary (i.e., temperature = 0.0); and \textbf{self-consistency} decoding \cite{wang_self-consistency_2022}, where the LM generates multiple reasoning chains and chooses the final chain based on majority voting on the evaluated answer (we use a temperature of 0.4 and 40 generations for all datasets).\footnote{Note that we do not report the performance of standard prompting with self-consistency decoding, since when the number of sampled outputs is large enough, this converges to standard prompting with greedy decoding \cite{wang_self-consistency_2022}.} We reproduce the baseline results ourselves in cases when they are not reported on certain tasks or when we clean the test set.

\subsection{LMs} We use OpenAI Codex \cite{chen_evaluating_2021} (\texttt{code-davinci-002}) in Section~\ref{sec:results} and experiment with four other code-generation LMs in Appendix~\ref{sec:analysis_LM_effect}.

\section{Results}
\label{sec:results}

\begin{table*}[!t]
\centering
\scalebox{0.7}{
\begin{tabular}{c|ccccc|c|ccc|c}
    \toprule
     & \multicolumn{5}{c|}{\textbf{Math Word Problems}} & \textbf{Planning} & \multicolumn{3}{c|}{\textbf{Multi-hop QA}} & \textbf{Relation} \\
     \textbf{Method} & GSM8K & SVAMP & MultiArith & ASDiv & AQuA & SayCan & StrategyQA & Date & Sport  & CLUTRR \\
     \hline
     
     \multicolumn{11}{c}{Greedy Decoding} \vspace{0.05cm} \\
     \hline
     Standard  & 19.6 & 69.5 & 43.8 & 72.1 & 31.5 & 82.5 & 63.9 & 51.3 & 71.9 & 42.0 \\
     CoT & 63.3 & 77.3 & 96.5 & 80.0 & 42.1 & 86.4 & \textbf{72.5} & 59.9 & 98.6 & 48.5  \\
     LtM &  38.3 & 80.3 & 74.0 & 76.5 & 40.6 & 77.7 & 72.2 & 76.6 & \textbf{99.5} & 47.2 \\ 
     Faithful CoT (ours)  & \textbf{72.3} & \textbf{83.4} & \textbf{98.8} & \textbf{80.2} & \textbf{47.2} & \textbf{89.3} & 63.0 & \textbf{81.6} & 99.1 & \textbf{58.9} \\
     \hline
     \multicolumn{11}{c}{Self-Consistency Decoding} \vspace{0.05cm} \\
     \hline
     CoT & 78.0 & 86.8 & \textbf{100.0} & 84.2 & 52.0 & 89.3 & \textbf{79.8} & 63.8 & 98.0 & 45.7 \\
     LtM &  38.8 & 80.5 & 74.0 & 76.3 & 44.9 & 76.7 & 71.9 & 77.2 & \textbf{99.4} & 50.9 \\ 
     Faithful CoT (ours) & \textbf{80.0} & \textbf{88.8} & 99.2 & \textbf{84.4} & \textbf{61.4} & \textbf{94.2} & 65.2 & \textbf{85.5} & 99.0 & \textbf{71.9} \\
     \bottomrule
\end{tabular}
}
\caption{Accuracy of different prompting methods on 10 reasoning datasets from 4 domains. We compare our method, Faithful CoT, with standard \cite{brown_language_2020}, CoT \cite{wei_chain_2022}, and Least-to-Most prompting \cite{zhou_least--most_2022}, with \texttt{code-davinci-002} as the LM.
The best results within each decoding strategy are \textbf{bolded}.
}
\label{tab:main}
\end{table*}

Our results on all datasets are shown in Table~\ref{tab:main}. With \texttt{code-davinci-002} as the Translator, Faithful CoT outperforms all baselines across the vast majority of datasets and domains under both decoding strategies. With \textbf{greedy decoding}, Faithful CoT outperforms all baselines on 8 of the 10 datasets, by a relative improvement of up to 14.2\% on MWP, 3.4\% on Planning, 6.5\% on Date Understanding from Multi-hop QA, and a surprising 21.4\% on Relational Inference. Generally, we see larger gains on harder datasets. Take MWP as an example: on simpler datasets where CoT already performs decently (e.g., MultiArith and AsDiv, where most questions require only 1-2 steps to solve), the gains are smaller (0.3\% to 2.4\%); however, we see the largest gain (14\%) on the most difficult GSM8K, which requires up to 8 steps to solve. With \textbf{self-consistency decoding}, Faithful CoT still performs the best on 7 out of the 10 datasets. Compared to greedy decoding, the relative gain increases on 4 datasets (AQUA: 12.1\% $\rightarrow$ 18.1\%, SayCan: 3.4\% $\rightarrow$ 5.5\%, Date Understanding: 6.5\% $\rightarrow$ 10.8\%, and CLUTRR: 21.4\% $\rightarrow$ 41.3\%), but decreases or remains unchanged for the remaining three MWP datasets (GSM8K: 9.0\% $\rightarrow$ 2.6\%, SVAMP: 3.9\% $\rightarrow$ 2.3\%, ASDiV 0.2\% $\rightarrow$ 0.2\%).

On the other hand, we do not see clear empirical gains on two multi-hop QA datasets, Sports Understanding on StrategyQA. On Sports Understanding, Faithful CoT and LtM both have near perfect accuracy (99+), suggesting that the dataset is almost saturated. On StrategyQA, however, the performance of our method is still far from the baselines. To understand why, we specifically compare the examples where CoT makes a correct prediction but our method fails. As shown in Figure~\ref{fig:strategyQA_errors} in Appendix~\ref{sec:appendix_error_analysis}, we find that the likely primary cause is the sparsity of Datalog in the pretraining data for Codex, as an overwhelming 29\% of errors are syntax-related. Moreover, including Datalog in the prompt also interferes with NL generation, making it harder for Codex to produce relevant subquestions (17\%), retrieve knowledge correctly (10\%), and come up with valid reasoning from the knowledge to the answer (10\%). Another potential cause is the nature of the task, as the difficulty for many StrategyQA questions does not lie in reasoning but rather in knowledge retrieval, which makes the advantages of our deterministic solver less obvious. Still, with further pretraining on Datalog, we believe that there is room for improvement.

To see how generalizable our method is, we also experiment with four alternative LMs and observe consistent gains brought by Faithful CoT over the baselines, as shown in Appendix~\ref{sec:analysis_LM_effect}. In particular, with GPT-4, we set the new few-shot SOTA results on 7 datasets, achieving 95.0+ accuracy in four out of five MWP and two out of three Multi-hop QA datasets. Overall, these results suggest that faithfulness does empirically improve performance.

\section{Analysis}
\label{sec:analysis} 

In this section, we perform an extensive analysis of the strengths and weaknesses of our method, to better understand the role of different components, the robustness to design choices, the plausibility of generated reasoning chains, as well as frequent error patterns where it still struggles. Here, we only show the first two aspects; see the rest in Appendix~\ref{sec:appendix_extended_analysis}. Unless otherwise stated, we choose one dataset from each domain (GSM8K, Date Understanding, SayCan, and CLUTRR) and use \texttt{code-davinci-002} outputs with greedy decoding.

\subsection{Ablation Study}
\label{sec:analysis_ablation}

\begin{figure}[t!]
\pgfplotsset{width=0.95\columnwidth, height=3cm}
    \centering
    \begin{tikzpicture}  
        \begin{axis}
        [  
            ybar,
            ymin=0, ymax=100,
            ytick={0, 20, 40, 60, 80, 100},
            major x tick style = transparent,
            bar width=4pt,
            enlarge x limits=0.25,
            ylabel={Accuracy},
            ylabel style={font=\small},
            symbolic x coords={GSM8K, Date, SayCan, CLUTRR},  
            xtick=data,
            xticklabel style={font=\small},
            yticklabel style={font=\scriptsize},
                axis x line*=bottom,
                axis y line*=left,
                legend style={draw=none},
        legend cell align=left,
                legend style={
                        at={(1.0,1.05)},
                        anchor=south east,
                        column sep=1ex,
                        font=\tiny,
                        legend columns=3
                },
            ]
        \addplot[ybar,  fill=GnBu-C, postaction={}] coordinates {
            (GSM8K, 72.3) (Date, 81.6) (SayCan, 89.3) (CLUTRR, 58.9)
        };
        \addplot[ybar,  fill=GnBu-E, postaction={pattern=north west lines}] coordinates {
            (GSM8K, 75.4) (Date, 83.0) (SayCan, 0.0) (CLUTRR, 51.8)
        };  
        \addplot[ybar, fill=GnBu-G, postaction={pattern=north east lines}] coordinates {
            (GSM8K, 73.5) (Date, 80.2) (SayCan, 0.0) (CLUTRR, 39.6)
        };  
        \addplot[ybar, fill=GnBu-I, postaction={pattern=dots}] coordinates {
            (GSM8K, 72.8) (Date, 79.7)  (SayCan, 90.3) (CLUTRR, 9.3)
        };  
        \addplot[ybar, fill=GnBu-K, postaction={}] coordinates {
            (GSM8K, 21.5) (Date, 57.9) (SayCan, 90.3) (CLUTRR, 40.9)
        };  
        \legend{Full, No rationale, No NL but nudge, No NL, No solver}  

        \end{axis}  
    \end{tikzpicture}
    \vspace{-0.1in}
    \caption{
    Ablation study results: accuracy when we remove different parts of the prompt. See Section~\ref{sec:analysis_ablation} for details.
    }
    \vspace{-0.15in}
    \label{fig:ablation}
\end{figure}

Given the strong performance of Faithful CoT, we now address a natural question: \textbf{how much does each part of the prompt contribute to the accuracy?} We perform an ablation study where we remove different parts of the prompt and see how the performance changes. In addition to the original prompt (``Full''), we test four variations, illustrated with the example from Figure~\ref{fig:baseline_and_our_prompts}:\\
\textbf{No rationale.} We remove the rationales, i.e., everything in the brackets from the NL comments, e.g., ``independent, support: [`There are 15 trees']''.\\
\textbf{No NL but nudge.} We remove all NL comments except the ``nudge'' line: e.g., ``\# To answer this question, we write a Python program to answer the following subquestions''.\\
\textbf{No NL.} We remove all NL comments.\\
\textbf{No solver.} Instead of calling the external solver, we add ``Answer: \{answer\}'' to the end of every exemplar and let the LM predict the answer itself.

Figure~\ref{fig:ablation} shows the results of all prompt variations. On GSM8K, Date Understanding, and SayCan, NL comments contribute little to the performance, and sometimes even slightly hurt it. On CLUTRR, however, their role is crucial, since the exclusion of each component (rationale, nudge, subquestions) results in a clear accuracy drop. In particular, comparing \textbf{No NL but nudge} and \textbf{No NL}, the nudge line itself brings a striking improvement by 31.3 points. 

The external solver relieves the burden of problem solving from the LM. Without it, the accuracy suffers a huge decline on GSM8K, Date Understanding, and CLUTRR (-50.8, -22.9, and -19.4 respectively), while on SayCan it improves by 2.9 nonetheless. One potential influencing factor is that SayCan might be too homogeneous, as it contains a set of only 3 predefined actions. This can make the task relatively easy, which allows all model variants to achieve around 90\% accuracy and renders the solver unnecessary. Another potential reason is the level of correspondence between the final answer and the reasoning chain for different datasets: as shown in Figure~\ref{fig:method}, the answer in SayCan is a sequence of actions (e.g., \texttt{find(redbull)}), each directly corresponding to one step in the reasoning chain (e.g., \texttt{at redbull trash}). However, the answer in the other three datasets is only a single number or string, which can only be derived after executing \textit{all} the steps in the reasoning chain. Therefore, the latter type of tasks further necessitates the presence of an external solver.

\subsection{Robustness to Exemplars}

\begin{table}[t!]
\centering
\scalebox{0.7}{
\begin{tabular}{c|rrrr}
    \toprule
     \textbf{Exemplars} & GSM8K & Date & SayCan & CLUTRR \\
     \hline
     Set 0 (Table~\ref{tab:main})  & 72.3 & 81.6 & 89.3 & 58.9  \\
     Set 1 & 72.6 & 81.3 & 91.3 & 59.0 \\
     Set 2 & 71.1 & 85.0 & 85.4 & 57.2 \\
     Set 3 & 72.3 & 82.5 & 88.3 & 58.0 \\
     Set 4 & 71.2 & 77.4 & 88.3 & 55.5 \\
     Set 5 & 71.5 & 85.0 & 89.3 & 56.0 \\
     \hline
     \textbf{Mean} & 71.8 & 82.1 & 88.7 & 57.4 \\
     \textbf{Std} & 0.6 & 2.8 & 1.9 & 1.5 \\
     \bottomrule
\end{tabular}
}
\caption{Robustness to the choice of exemplars.}
\vspace{-0.2in}
\label{tab:robustness}
\end{table}

We now answer the next question: \textbf{how much does the choice of exemplars matter}? To do this, we annotate 20 examples in total, randomly sample k (7-10, depending on the dataset) to construct the prompt, and repeat the process five times. Table~\ref{tab:robustness} shows the performance of all six runs, including the original (from Table~\ref{tab:main}). The mean accuracy is close to the original (-1.5 to +1.2), still above the baselines by a large margin (7 to 17) on all datasets except the arguably easiest SayCan, considering the standard deviation (1.3 to 2.9). This strongly suggests that the benefits of Faithful CoT are minimally influenced by the choice of exemplars.

\subsection{Human Evaluation of Plausibility}
\label{sec:human_eval_details}

Our main experiments use final answer accuracy as the performance measure, but this does not necessarily correspond to the validity of the reasoning chain. Technically, a model can sometimes accidentally arrive at the correct answer with an invalid reasoning chain.
We then ask: \textbf{when the answer is correct, how often is the reasoning chain truly correct?} In other words, we want to evaluate the \emph{plausibility} of the reasoning chains. 

We conduct a human evaluation study on Prolific: given a generated reasoning chain that results in a correct answer, a crowd-worker selects whether it is \textit{A) completely correct}, or, if incorrect, specify why with \textit{B) incorrect NL} and/or \textit{C) incorrect SL}. Alternatively, they can select \textit{D) flawed question} and \textit{E) I am confused}.\footnote{See Appendix~\ref{sec:appendix_human_eval_details} for more details on the human study.}

\begin{figure}
    \centering
    \includegraphics[width=0.99\linewidth]{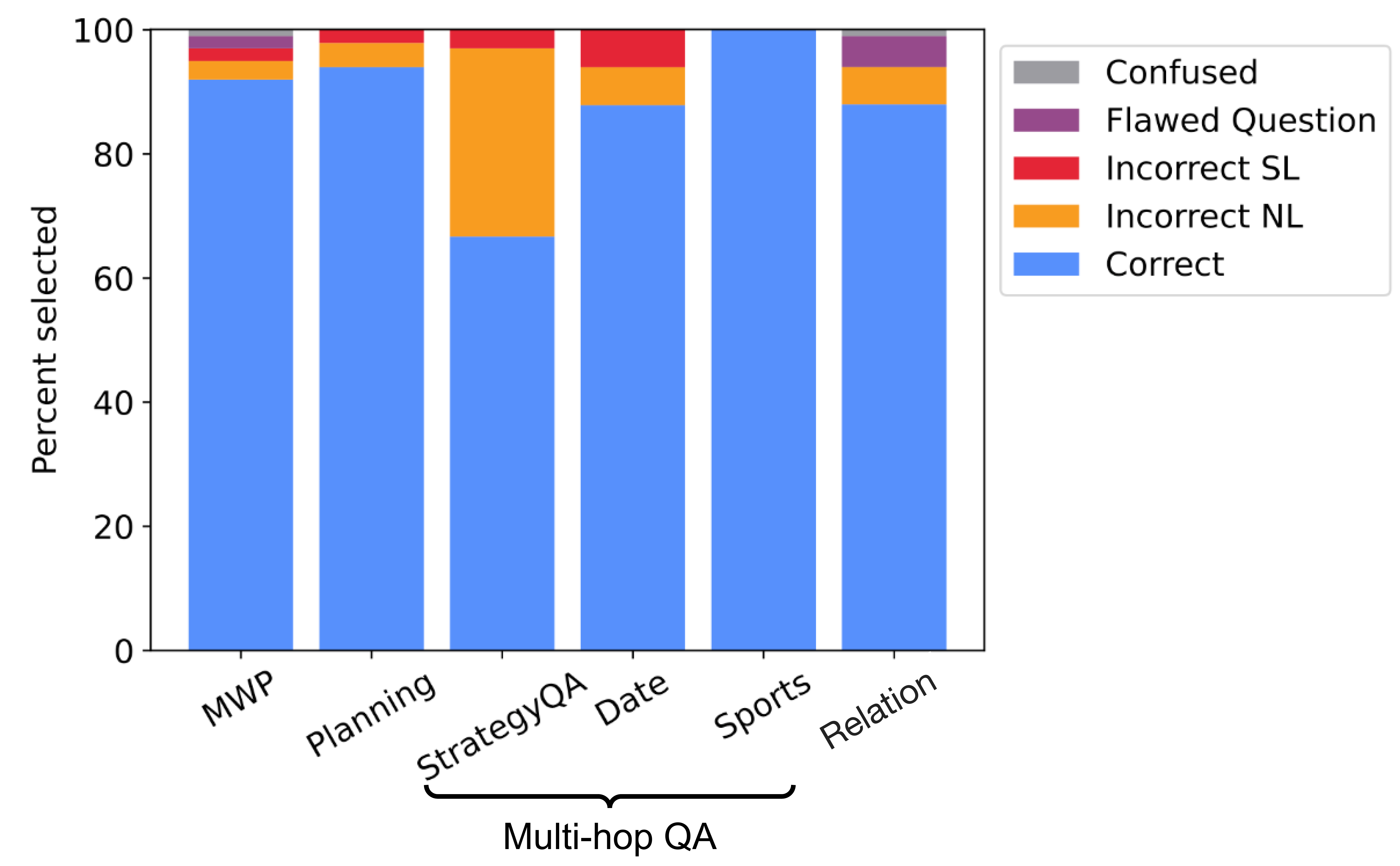}
    \caption{Human evaluation results of reasoning chain plausibility. Each column represents the percent of different answer choices selected by human evaluators in each domain/dataset.}
    \label{fig:human_eval}
\end{figure}

The results of our study are shown in Figure~\ref{fig:human_eval} (see Table~\ref{tab:human_eval} in the Appendix for numerical results). For most domains, we see that annotators often find the reasoning chain fully correct -- Sports, SayCan, and MWP have a 90\%+ correctness rate. To gain more insight into when the reasoning chain can be "incorrect", we perform an in-depth analysis of user annotations for the three worst-performing datasets -- StrategyQA (66.7\% correctness), Date (87.9\% correctness), and CLUTRR (88.0\% correctness). From our inspection, we find that annotators mistakenly mark a correct reasoning chain as incorrect at different rates based on the task (8.3\% of the time for StrategyQA, 41.7\% for Date Understanding, and 100\% for CLUTRR). We find annotators are inaccurate for Date because they incorrectly believe the generated code misuses a Python library (\texttt{relativedelta}), or they complain that there is a better way to answer the question. For CLUTRR, annotators mark chains as incorrect due to known ambiguity in the dataset. For example, the grandmother of one's child may not necessarily be their parent, but also a parent-in-law.

As for the remaining truly incorrect reasoning chains, we find the LM can sometimes add unnecessary steps or arrive at the correct answer by chance. The latter is especially an issue in StrategyQA - given that all questions have a True/False answer, it is common for an incorrect reasoning chain to result in a correct answer. For example, the LM correctly answers the question "Was Karachi a part of Alexander the Great's success?" as "True." However, the reasoning chain contains the flawed subquestion "Which countries are in Pakistan? Pakistan includes Pakistan, Afghanistan, and India." Though the final answer is correct and faithful to the LM generation, the reasoning chain contains wrong knowledge.

Overall, Faithful CoT does generate valid reasoning chains for the vast majority of the time when the answer is correct. However, we still see exceptions where the model arrives at the right answer via an incorrect reasoning chain. Though this happens infrequently, it raises concerns about when people should trust LMs. To our knowledge, we are the first to conduct a systematic human study on the plausibility of CoT-style reasoning chains, and we hope to see future work further investigate and improve on the flaws that our study brings to light.

\section{Conclusion}

We propose Faithful CoT, a framework that decomposes complex reasoning into Translation and Problem Solving. 
It guarantees that the reasoning chain is a faithful explanation of how the model arrives at the answer. We demonstrate the efficacy of our approach on 4 types of complex reasoning problems: Math Word Problems, Multi-hop QA, Planning, and Relational Inference. Our method sets new SOTA performance on 7 of the 10 datasets, while additionally providing a faithful explanation for the final answer. These results give empirical evidence that improving model \textit{interpretability}, by guaranteeing the faithfulness of an explanation, does not come at the expense of overall \textit{performance}; in fact, we see a strong synergy in between. Through a comprehensive analysis of the strengths and weaknesses of our method, we show its robustness to the choice of exemplars, the pivotal role of the solver, as well as frequent error patterns where it still struggles.

\section*{Limitations}
\label{sec:limitations}

One crucial limitation of our study is that on March 23rd, 2023, OpenAI discontinued the use of code-davinci-002. This has rendered part of our results unreplicable for any teams or researchers who have not been granted continued access to the model. This discontinuation was unexpected during our study. It raises important questions about using closed-source models for academic research.

Meanwhile, one methodological limitation of our approach lies in the scope of faithfulness. Currently, we guarantee that Problem Solving stage is faithful. However, the Translation stage is still opaque, meaning it is not self-interpretable how the LM generates the reasoning chain from the question. It is still an under-explored question whether it is possible to improve the interpretability of the LM generation process in general, and a few recent studies have made promising early progress \cite{yin2022interpreting, sarti2023inseq} that might be used to improve the faithfulness of the Translation stage. 

Finally, it still needs further exploration of the role NL comments in the reasoning chain. From our ablation study, in terms of \textit{performance}, whether to include the NL comments in the reasoning chain does not make a big difference on many of the datasets, especially those where the task is not knowledge-intensive. Nevertheless, speaking of \textit{interpretability}, NL comments can make the reasoning chain more structured and understandable to the end user. Further, NL comments can be an interface that allows users \textit{without} a programming background to interact with and debug the model, which we leave for future work.

\section*{Ethics Statement}
With the recent success of generative large LMs, they are now being used to solve complex reasoning problems. When using the output of an LM for reasoning, there is a danger that if the reasoning {\em appears} realistic, then the final answer or conclusion will also be considered reliable. As we highlighted in Figure~\ref{fig:cot_unfaithful} and ~\ref{fig:cot_unfaithful_full}, this is often not true, since an LM may produce a reasoning chain that looks plausible, but the final answer is still wrong. This work is a step in the direction of making the use of LMs more trustworthy by using the LM for just expressing its reasoning in a symbolic program and executing the program independently. In this work, we have ensured the faithfulness of the reasoning chain w.r.t how the final answer is produced in a variety of domains, but admittedly the Translation phase is still opaque. Therefore, our pipeline is still not entirely interpretable. Furthermore, as we have stressed in Section~\ref{sec:intro}, faithfulness does not guarantee correctness, so our method can still sometimes produce erroneous answers, which may pose a risk for users that rely on it for decision making.

\section*{Acknowledgements}
This research is based upon work supported in part by the DARPA KAIROS Program (contract FA8750-19-2-1004), the DARPA LwLL Program (contract FA8750-19-2-0201), the IARPA HIATUS Program (contract 2022-22072200005), and the NSF (Award 1928631). Approved for Public Release, Distribution Unlimited. The views and conclusions contained herein are those of the authors and should not be interpreted as necessarily representing the official policies, either expressed or implied, of DARPA, IARPA, NSF, or the U.S. Government.

We appreciate the support from OpenAI on increasing the rate limit for the Codex API. We also thank Jiani Huang, Ziyang Li, Litao Yan, Andrew Head, Mayur Naik, and Lyle Ungar for their valuable feedback.

\bibliography{custom}
\bibliographystyle{acl_natbib}


\newpage
\appendix

\section{Implementation Details}
\label{sec:appendix_implementation}

In all our experiments, we use OpenAI GPT-3 (\texttt{text-davinci-001} and \texttt{text-davinci-002}) and Codex  (\texttt{code-davinci-001} and \texttt{code-davinci-002}) models through the Python API available at \url{beta.openai.com}, from Sept, 2022 to Jan, 2023. The inference cost per example is \$0 for all Codex models since they are in limited beta period, and \$0.01 - \$0.03 for GPT-3 models depending on the dataset. It takes 2-15 seconds to run inference on one example with Codex models under a rate limit of 150,000 tokens/minute, and 1-8 seconds with GPT-3 models under 250,000 tokens/minute, also depending on the dataset. For example, on the GSM8K test set of 1,319 examples, it takes 3.5h to finish the inference with Codex and 2.3h with GPT-3.

We use the following hyper-parameters throughout all experiments:

\textbf{temperature}: 0.0 for greedy decoding, 0.4 for self-consistent decoding;\\
\textbf{max\_tokens}: 1000;\\
\textbf{n}: 1 for greedy decoding, 40 for self-consistent decoding;\\
\textbf{frequency\_penalty}: 0;\\
\textbf{presence\_penalty}: 0.

Any unspecified hyper-parameters are set to the default value on \url{https://beta.openai.com/docs/api-reference/completions/create}.

\section{Extended Results}
\label{sec:appendix_extended_results}

In this section, we present more results that do not fit into the main text.

\subsection{Examples of Unfaithful CoT Output}
\label{appendix:unfaithful_output}

\begin{figure}[h]
\centering
\includegraphics[width=\columnwidth]{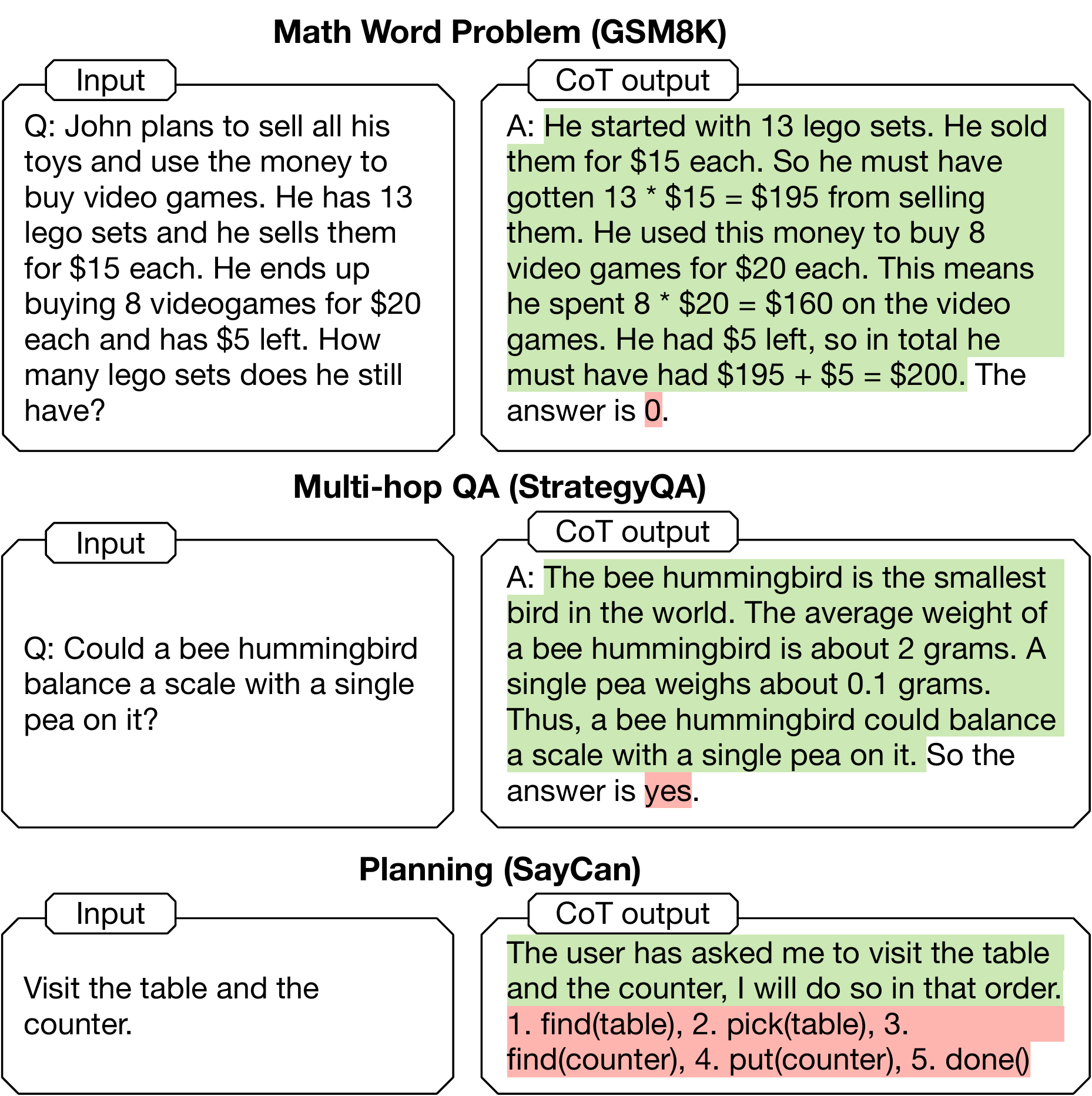}
\caption{Examples of \textit{unfaithful} output from CoT prompting \cite{wei_chain_2022} on three datasets. The answer (green) does not follow from the reasoning chain (blue). }
\label{fig:cot_unfaithful_full}
\end{figure}

Figure~\ref{fig:cot_unfaithful_full} shows examples of unfaithful output from the CoT method \cite{wei_chain_2022} on three datasets: GSM8K (MWP), StrategyQA (Multi-hop QA), and SayCan (Planning). Note that here we only show model outputs that are apparently unfaithful. In other cases, there is no \textit{known} inconsistency between the reasoning chain and the answer, but this still does not imply that the reasoning chain is a faithful explanation, since the ground truth (how the model predicts the answer) is unknown.

The GSM8K example is explained in Section~\ref{sec:intro}. In the StrategyQA example, though the reasoning chain correctly identifies that a hummingbird weighs much more than a pea, the answer is still ``yes''; in the SayCan example, the reasoning chain only mentions ``visit the table and the counter'', but the plan contains unnecessary ``pick'' and ``put'' operations.

 \citet{wei_chain_2022} claim that CoT ``provides an interpretable window into the behavior of the model, suggesting how it might have arrived at a particular answer and providing opportunities to debug where the reasoning path went wrong". As we have pointed out in Section~\ref{sec:intro}, since CoT does not guarantee faithfulness, how the model arrives at the answer could differ drastically from what is shown in the reasoning chain. Furthermore, it is still hard for the user to debug the model: even if they manually correct the reasoning chain and let the model regenerate the answer, it might still be wrong, since there is no causality between the reasoning chain and the answer.

\subsection{Comparison with Few-shot SOTA}
\label{sec:appendix_sota}

\begin{table*}[!t]
\centering
\scalebox{0.7}{
\begin{tabular}{c|rrrrr|r|rrr|r}
    \toprule
     & \multicolumn{5}{c|}{\textbf{Math Word Problems}} & \textbf{Planning} & \multicolumn{3}{c|}{\textbf{Multi-hop QA}} & \textbf{Relation} \\
     \textbf{Method} & GSM8K & SVAMP & MultiArith & ASDiv & AQuA & SayCan & StrategyQA & Date & Sport  & CLUTRR \\
     \hline
     Few-shot SOTA & 92.0 & 89.1 & \textbf{100.0} & 87.8 & \textbf{76.4} & 88.3 & \textbf{81.6} & 76.2 & 98.5 & 50.9 \\
     Faithful CoT (ours) &  \textbf{95.0} & \textbf{95.4} & 99.2 & \textbf{95.6} & 73.6 & \textbf{93.2} & 65.2 & \textbf{95.8} & \textbf{99.3} & \textbf{71.9} \\ 
     $\delta_{acc}$ & \textcolor{ForestGreen}{+3.0} & \textcolor{ForestGreen}{+6.3} & \textcolor{RubineRed}{-0.8} & \textcolor{ForestGreen}{+7.8} & \textcolor{RubineRed}{-2.8} & \textcolor{ForestGreen}{+4.9} & \textcolor{RubineRed}{-16.4} & \textcolor{ForestGreen}{+19.6} & \textcolor{ForestGreen}{+0.8} & \textcolor{ForestGreen}{+21.0} \\
     \bottomrule
\end{tabular}
}
\caption{Comparison between the existing few-shot SOTA results and the optimal Faithful CoT results (with the best-performing LM (\texttt{code-davinci-002} for SayCan and CLUTRR, and \texttt{gpt-4} for the rest of the datasets). See Appendix~\ref{sec:appendix_sota} for sources of SOTA results.}
\label{tab:SOTA}
\end{table*}


We compare the results of Faithful CoT with the published few-shot SOTA in Table~\ref{tab:SOTA}. The Faithful CoT results are obtained with the best-performing LM among all five LMs we experiment with (see Section~\ref{sec:analysis_LM_effect} on each dataset. The SOTA results are obtained from the following sources:

\textbf{GSM8K}: \citet{openai2023gpt4};

\textbf{MultiArith, ASDiv, StrategyQA}: \citet{wang_self-consistency_2022};

\textbf{SVAMP}: \citet{chen_program_2022};

\textbf{AQuA}: \citet{pitis2023boosted};

\textbf{SayCan, Sports Understanding}: \cite{wei_chain_2022};

\textbf{Date Understanding}: \cite{gao_pal_2022};

\textbf{CLUTRR}: Our implementation of LtM prompting \cite{zhou_least--most_2022} (no existing work reports few-shot performance on CLUTRR with $K$ up to 10).

With Codex and GPT-4, Faithful CoT sets new SOTA performance on 7 out of the 10 datasets across four domains, achieving 95.0+ accuracy on 6 of them.

\subsection{Empirical Comparison with Concurrent Work}
\label{sec:appendix_concurrent_work}



\begin{figure}
    \centering
    \includegraphics[width=\columnwidth]{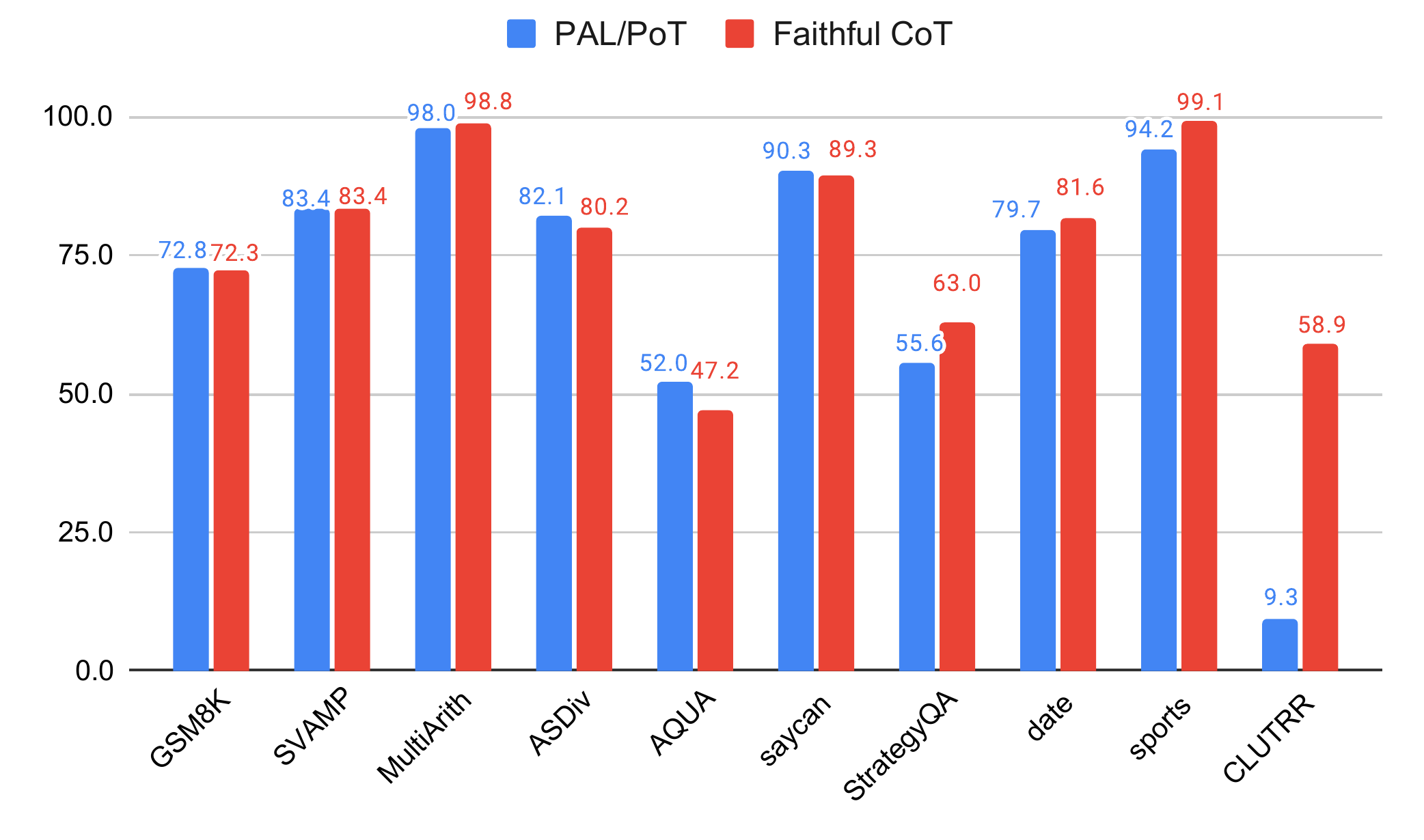}
    \caption{Accuracy of our method and two concurrent methods, Program of Thoughts (POT) \cite{chen_program_2022} and Program-Aided Language Models (PAL) \cite{gao_pal_2022}, on 10 reasoning datasets.}
    \label{fig:fcot_vs_pal_pot}
\end{figure}

Two pieces of concurrent work, Program of Thoughts (PoT) \cite{chen_program_2022} and Program-Aided Language Models (PAL) \cite{gao_pal_2022}, were announced on arXiv within three months of our work. Essentially, they both generate Python programs, or SL-only reasoning chains, to derive the answer. Our approach differs from them mainly in the additional component of structured NL comments, which decomposes the original problem into simpler, inter-dependent subproblems. 

Aside from the qualitative differences highlighted in Section~\ref{sec:related_work}, we perform an empirical performance comparison with them on the same set of 10 datasets used in our main evaluation. Since both papers have only tackled math reasoning and symbolic reasoning tasks, we reimplement their methods by using the ``noNL'' prompt in our ablation study from Section~\ref{sec:analysis_ablation}. The comparison is done with \texttt{code-davinci-002} as the underlying LM and greedy decoding.

As shown in Figure~\ref{fig:fcot_vs_pal_pot}, on 6 of the 10 datasets (including most MWP datasets, SayCan, and Date Understanding), PAL/PoT and Faithful CoT have very close accuracy (<2.0 difference). On AQuA, PAL/PoT is visibly better. On the remaining three datasets (StrategyQA, Sports Understanding, and CLUTRR), Faithful CoT reasonably outperforms PAL/PoT. This may suggest that our method has an advantage when the task requires extensive external knowledge (e.g., StrategyQA and Sports Understanding) or when the SL is not frequent in the LM's pretraining data (e.g., Datalog, or our self-defined relational expressions). 

Finally, note that the key contribution of our method lies in interpretability. Though the addition of structured NL comments sometimes does not make a difference in performance, it does make the reasoning chain more understandable to the user. Furthermore, it may even enable users without a programming background to debug the model, by only interacting with the NL subproblems (e.g., adding/removing/editing a subproblem), which is worth further exploration in the future.
 
\section{Extended Analysis}
\label{sec:appendix_extended_analysis}

\subsection{Ablation Study}

Table~\ref{tab:appendix_ablation} shows the full results of the ablation study from Section~\ref{sec:analysis_ablation}.

\begin{table}[h!]
\centering
\vskip 0.15in
\scalebox{0.8}{
\begin{tabular}{l|rrrr}
    \hline
     \textbf{Prompt} & GSM8K & Date & SayCan & CLUTRR \\
     \hline
    Full & 72.3 & 81.6 & 89.3 & \textbf{58.9} \\ 
    No rationale & \textbf{75.4} & \textbf{83.0} & - & 51.8 \\ 
    No NL but nudge  & 73.5 & 80.2 & - & 39.6 \\ 
    No NL & 72.8 & 79.7 & 90.3 & 9.3 \\ 
    No solver & 21.5 & 57.9 & \textbf{90.3} & 40.9 \\
     \hline
\end{tabular}
}
\caption{Ablation study results that accompany  Figure~\ref{fig:ablation}. We report accuracy when we remove different parts of the prompt.}
\label{tab:appendix_ablation}
\end{table}

\subsection{Robustness to Prompt Phrasing}
\label{sec:appendix_robustness_to_prompt_phrasing}

\begin{table}[h!]
\centering
\vskip 0.15in
\scalebox{0.8}{
\begin{tabular}{l|rrrr}
    \hline
     \textbf{Prompt} & GSM8K & Date & SayCan & CLUTRR \\
     \hline
     Original  & 72.3 & 81.6 & 89.3 & 58.9  \\
     Variation 1 &  69.1 & 84.4 & 88.3 & - \\ 
     Variation 2 & 70.3 & 81.6 & 90.3 & 56.2  \\
     Variation 3 & 70.2 & 80.5 & 87.4 & 55.9 \\
     \hline
     \textbf{Mean} & 70.5 & 82.0 & 88.8 & 57.0 \\
      \textbf{Std} & 1.3 & 1.7 & 1.3 & 1.7 \\

     \hline
\end{tabular}
}
\caption{Robustness to prompt phrasing.}
\label{tab:prompt_phrasing}
\end{table}

We study the sensitivity of our method to subtle differences in the prompt design. We experiment with three prompt variations: 1. randomly permuting the order of independent subquestions/reasoning steps; 2: Changing the variable names; 3. changing the nudge line (e.g. from ``\# To answer this question, write a Python program to answer the following subquestions'' to ``\# To solve this question, we answer each of the following subquestions with a Python program'').

We rerun the evaluation of all three variations on 4 datasets (when applicable) used in the Section~\ref{sec:analysis}, under greedy decoding. Table~\ref{tab:prompt_phrasing} shows the results. Overall, the performance is quite stable, always above each baseline on all four datasets.

\subsection{Model Sensitivity}
\label{sec:analysis_LM_effect}

In this section, we want to answer the question: \textbf{how much does the choice of LM matter?} All results in Section~\ref{sec:results} are obtained using \texttt{code-davinci-002}. Here, we examine the effect of using four alternative code-generation models as the Translator: \texttt{code-davinci-001}, \texttt{text-davinci-002}, \texttt{text-davinci-003}, \texttt{gpt-4}. We compare our method with the three baselines using each of the above LM on five MWP datasets, using the greedy decoding strategy.

As shown in Table~\ref{tab:model_sensitivity}, regardless of the underlying LM, Faithful CoT consistently outperforms all baselines on the vast majority of the datasets, and performs very closely with the best method (<2.0 difference) on the remaining ones. On average, it has a relative accuracy gain of 16.1\%, 11.0\%, 9.4\%, and 4.6 \% over the best-performing method among the baselines, for each LM respectively. This indicates that even though the absolute performance varies depending on the LM, Faithful CoT brings a relatively consistent accuracy gain.

Notably, with GPT-4 as the underlying LM, Faithful CoT results in 95.0+ accuracy in 4 of the 5 MWP datasets, far outperforming the previous few-shot SOTA on three of them (GSM8K, SVAMP, and ASDiv).

\begin{table*}[h]
\centering
\scalebox{0.7}{
\begin{tabular}{c|rrrrr|r}
    \toprule
     \textbf{Method / Dataset} & GSM8K & SVAMP & MultiArith & ASDiv & AQuA & Average \\
     \hline
     & \multicolumn{5}{c}{LM: \texttt{code-davinci-001}} \vspace{0.05cm} \\
     \hline
     Standard & 5.2 & 28.7 & 8.6 & 38.5 & \textbf{22.8} & 20.8 \\ 
     CoT & 14.7 & 41.2 & 57.2 & 50.4 & 22.4 & 37.2 \\ 
     LtM &  9.8 & 44.9 & 18.5 & 46.6 & 20.5 & 28.1 \\ 
     Faithful CoT (ours) & \textbf{27.4} & \textbf{50.8} & \textbf{63.3} & \textbf{53.7} & 20.9 & \textbf{43.2} \\

     \hline
     & \multicolumn{5}{c}{LM: \texttt{text-davinci-002}} \vspace{0.05cm} \\
     \hline
     Standard &  15.4 & 65.2 & 34.1 & 64.8 & 24.0 & 40.7 \\ 
     CoT &  47.2 & 68.0 & 91.1 & 69.9 & 40.6 & 63.4 \\ 
     LtM &  32.9 & 73.8 & 68.3 & 70.9 & 34.6 & 56.1 \\ 
     Faithful CoT (ours) &  \textbf{62.7} & \textbf{80.0} & \textbf{92.8} & \textbf{75.4} & \textbf{41.3} & \textbf{70.4} \\

     \hline
     & \multicolumn{5}{c}{LM: \texttt{text-davinci-003}} \vspace{0.05cm} \\
     \hline
     Standard &  16.9 & 69.4 & 38.8 & 59.1 & 29.5 & 42.7 \\ 
     CoT &  59.6 & 79.5 & \textbf{95.0} & 69.1 & 46.9 & 70.0 \\ 
     LtM &  34.6 & 79.5 & 73.7 & 70.0 & 44.5 & 60.5 \\ 
     Faithful CoT (ours) & \textbf{71.7} & \textbf{85.1} & 94.5 & \textbf{80.7} & \textbf{50.8} & \textbf{76.6} \\

     \hline
     & \multicolumn{5}{c}{LM: \texttt{gpt-4}} \vspace{0.05cm} \\
     \hline
     Standard & 46.9 & 88.4 & \textbf{98.7} & 70.2 & 50.4 & 70.9 \\ 
     CoT &  64.9 & 80.0 & 94.0 & 71.6 & \textbf{75.2} & 77.1  \\ 
     LtM &  91.8 & 92.9 & 98.3 & 86.7 & 72.0 & 87.5 \\ 
     Faithful CoT (ours) &  \textbf{95.0} & \textbf{95.4} & 98.5 & \textbf{95.6} & 73.6 & \textbf{91.6} \\ 

     \bottomrule
\end{tabular}
}
\caption{Accuracy of different prompting methods with each underlying LM on 5 MWP reasoning datasets.}
\label{tab:model_sensitivity}
\end{table*}

\subsection{Enforcing Constraints}
\label{sec:appendix_constraint}

Since our generated reasoning chain contains structured components (e.g., dependency graphs), another natural question to ask is: \textbf{will it be helpful to enforce certain constraints on the generation?} Using MWP datasets as a case study, we examine the effect of three such constraints:

\textbf{Graph validity.} The dependency graph must be a Directed Cyclic Graph (DAG), e.g., it is not allowed for a subquestion to depend on itself.

\textbf{No over-dependency.} The code cannot depend on \textit{any} variable that its corresponding subquestion has not mentioned, e.g. in Figure~\ref{fig:baseline_and_our_prompts}, since Q5 says ``depend on 4'', then the corresponding code should \textit{not} use the variable \texttt{eggs\_in\_dozen}, since it is not the output of Q4.

\textbf{No under-dependency.} The code must depend on \textit{all} variables that its corresponding subquestion has mentioned, e.g. in the same example, since Q5 says ``depend on 4'', then the corresponding code \textit{must} use the variable \texttt{eggs\_in\_dozen}.

\begin{table}[t!]
\centering
\vskip 0.15in
\scalebox{0.7}{
\begin{tabular}{l|rrrrr}
    \hline
     \textbf{Constraint} & GSM8K & SVAMP & MultiArith & ASDiv & AQuA \\
     \hline
     None & 80.0  & 88.8  & 99.2  & 84.4  & 61.4  \\
     \hline
     + G & 0.0 & 0.0 & 0.0 & \color{OrRd-J}{-0.1} & \color{OrRd-J}{-0.8} \\
+ O & \color{OrRd-J}{-0.9} & \color{OrRd-J}{-0.1} & \color{OrRd-J}{-0.1} & \color{Greens-H}{+0.4} & \color{OrRd-J}{-3.9} \\
+ U & \color{OrRd-J}{-1.0} & \color{OrRd-J}{-3.6} & 0.0 & \color{OrRd-J}{-1.2} & \color{Greens-H}{+1.2} \\
+ GO & \color{OrRd-J}{-1.7} & \color{OrRd-J}{-0.4} & \color{OrRd-J}{-0.1} & \color{Greens-H}{+0.2} & \color{OrRd-J}{-3.9} \\
+ GU & \color{OrRd-J}{-1.0} & \color{OrRd-J}{-3.7} & 0.0 & \color{OrRd-J}{-1.2} & \color{Greens-H}{+0.8} \\
+ OU & \color{OrRd-J}{-4.0} & \color{OrRd-J}{-5.4} & \color{OrRd-J}{-0.1} & \color{OrRd-J}{-2.6} & \color{OrRd-J}{-4.3} \\
+ GOU & \color{OrRd-J}{-5.0} & \color{OrRd-J}{-5.9} & \color{OrRd-J}{-0.1} & \color{OrRd-J}{-3.2} & \color{OrRd-J}{-5.5} \\
     \hline
\end{tabular}
}
\caption{Accuracy change after enforcing different constraints on the generation. The ``None'' row shows the original performance without any constraint (from Table~\ref{tab:main}). Each row below adds a different set of constraints: G stands for ``graph validity'', O for ``no over-dependency'', and U for ``no under-dependency''. Results are on all MWP datasets under self-consistent decoding.}

\label{tab:constraint}

\end{table}



We investigate the effect of adding constraints on the generations under self-consistent decoding. Starting with our original results (without any constraint), we add a different set of constraints at each time and report the accuracy change in Table~\ref{tab:constraint}. Individually, the graph validity constraint results in little to no change in the performance, but the other two constraints lead to a more unstable change--mostly a decrease--across datasets. Adding two or more constraints further lowers the performance in almost all cases except on MultiArith (the easiest dataset), revealing the tradeoff between accuracy and satisfying the constraints. It also indicates that a proportion of generations (1.0\% to 8.9\%) in our existing results do not satisfy all constraints. However, it may still be worth enforcing some of these constraints (e.g., graph validity) at the cost of performance, in order for users to better control and interact with the model.  

\section{Human Evaluation Details}
\label{sec:appendix_human_eval_details}

\begin{table*}[h]
    \centering
    \vskip 0.15in
    \small
    \begin{tabular}{lrrrrr|r}
         \toprule
         Domain &  Correct & Incorrect NL & Incorrect SL & Flawed Question & Confused & Agreement\\
         \midrule
         MWP & 92.0 & 3.0 & 2.0 & 2.0 & 1.0 & 0.790\\
         Planning & 94.0 & 3.9 & 2.1 & 0.0 & 0.0 & 0.947\\         
         StrategyQA & 66.7 & 30.3 & 3.0 & 0.0 & 0.0 & 0.455\\         
         Date & 87.9 & 6.1 & 6.1 & 0.0 & 0.0 & 0.788\\
         Sports & 100.0 & 0.0 & 0.0 & 0.0 & 0.0 & 0.899\\         
         Relation & 88.0 & 6.0 & 0.0 & 5.0 & 1.0 & 0.683\\
         \bottomrule
    \end{tabular}
\caption{Numerical results for human evaluation of reasoning chain correctness, accompanying Figure~\ref{fig:human_eval}. Each row represents the percent (0-100) of different answer choices selected by human evaluators in each domain/dataset, as well as the inter-annotator agreement.}
    \label{tab:human_eval}
\end{table*}

We hire crowd workers on Prolific to evaluate the correctness of model-generated reasoning chains that result in a correct answer. We sample 100 reasoning chains for each domain generated by \texttt{code-davinci-002} with the greedy decoding strategy, where each set of 100 contains an equal number of samples from all datasets within the domain. We further require annotators to have experience coding in the programming language of the dataset they annotate (Python for MWP/CLUTRR/Sports/Date, and Scala for StrategyQA, as Datalog was not an option in Prolific). We have a single survey for each domain, with the exception of Multi-hop QA (in this case, we have separate surveys for StrategyQA, Date, and Sports, given the different nature of each dataset). Additionally, there was no way to ensure annotators knew PPDL on Prolific. In order to ensure high-quality annotations for SayCan, the authors annotate this dataset themselves. 

\begin{figure}
    \centering
    \includegraphics[width=\linewidth]{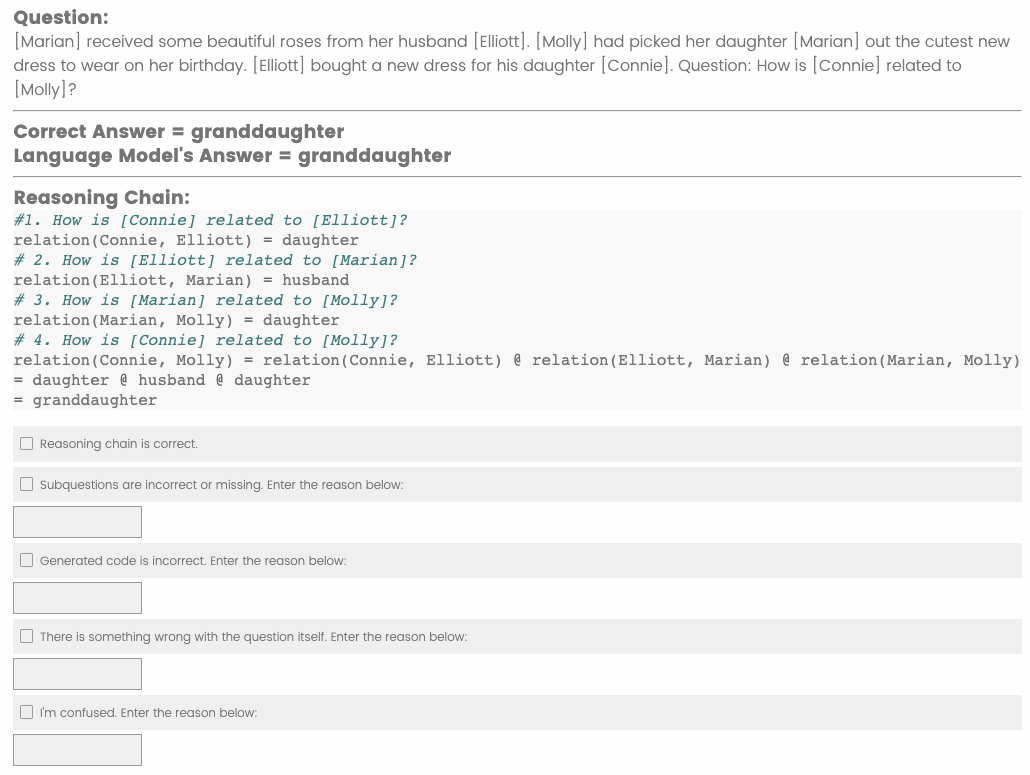}
    \caption{Example of our annotation interface for the CLUTRR survey}
    \label{fig:interface}
\end{figure}

\paragraph{Annotation Process} Each of the 100 questions for each domain is annotated by at least three annotators. Each annotator is given 10 questions to annotate. A screenshot of our survey interface is shown in Figure~\ref{fig:interface}. At the start of the survey, annotators are given examples of a correct reasoning chain, a reasoning chain with incorrect NL, and a reasoning chain with incorrect SL. The examples match the domain of reasoning chains in each survey. As an attention and understanding check, we ask annotators to label an example they had just seen. If the annotator fails this question, they are sent to the end of the survey and their responses are filtered out. We also manually filter out all spammers (annotators who answer with the same response repeatedly or complete the survey in under 3 minutes). After the surveys are complete, we compute annotator agreement and then take the majority label for each reasoning chain as the final label in our analysis.
Our annotator population consists of 100 annotators with an average age of 25 years and an average income of 40k per year. 87.9\% of the annotators are males and 56\% have a four year college degree.
Annotators are compensated at \$16/hr, and average 2 minutes per question. Our full study cost \$280. Sample instructions for the CLUTRR survey can be found in the Supplementary Materials.

\section{Dataset Details}
\label{sec:appendix_dataset_details}

\subsection{Statistics}

We show the dataset details in Table~\ref{table:datasets}, including the statistics, the number of few-shot exemplars used in the prompt, and example inputs and outputs.

In particular, we notice that in one of our baselines \citet{wei_chain_2022}, the reported number of exemplars used in the prompt is inconsistent between the main text (10) and the appendix (6). To ensure fair comparison, we rerun the baseline with 10 exemplars for our results in Table~\ref{tab:main}, which is what we use for our method. 

\begin{table*}[h]
\centering
\scalebox{0.78}{
\begin{tabular}{p{1.5cm}p{1.8CM}>{\raggedleft\arraybackslash}p{1.1cm}>{\raggedleft\arraybackslash}p{1cm}p{12.5cm}}
    \toprule \textbf{Domain} & \textbf{Dataset} & \bf{\# Shot}  & \bf{\# Test} & \textbf{Example}  \\
    \midrule \multirow{16}{1.5cm}{Math Word Problems} & GSM8K & 8 & 1,319 & Q: Natalia sold clips to 48 of her friends in April, and then she sold half as many clips in May. How many clips did Natalia sell altogether in April and May? \newline A: \texttt{72} \\
    & SVAMP & 8 & 1,000 & Q: Each pack of dvds costs 76 dollars. If there is a discount of 25 dollars on each pack. How much do you have to pay to buy each pack? \newline A: \texttt{51} \\
    & MultiArith & 8 & 600 &  Q: For Halloween Debby and her sister combined the candy they received. Debby had 32 pieces of candy while her sister had 42. If they ate 35 pieces the first night, how many pieces do they have left? \newline A: \texttt{39} \\
    & ASDiv & 8 & 2,096 &  Q: Seven red apples and two green apples are in the basket. How many apples are in the basket? \newline A: \texttt{9} \\
    & AQuA & 8 & 254 &  Q: A car finishes a journey in 20 hours at the speed of 60 km/hr. If the same distance is to be covered in 10 hours, how much speed does the car gain? \newline A: \texttt{``120 kmph''}\\
    \midrule
    \multirow{6}{1.5cm}{Multi-hop \newline QA} & StrategyQA & 6 & 2,290 & Q: Did Aristotle use a laptop? \newline A: \texttt{False} \\
    & Date \newline Understanding & 10 & 359 & Q: Yesterday was April 30, 2021. What is the date tomorrow in MM/DD/YYYY? \newline A: \texttt{``05/02/2021''} \\
    & Sports \newline Understanding & 10 & 977 &  Q: Is the following sentence plausible: ``Lebron James hit the turnaround jumper''? \newline A: \texttt{True} \\
    \midrule
    \multirow{3}{1.5cm}{Planning} & SayCan & 7 & 103 & Q: Could you get me a drink with caffeine? \newline A: \texttt{``1.find(redbull)\ 2.pick(redbull)\ 3.find(user)\ 4.put(redbull)\ 5.done().''} \\
    \midrule
    \multirow{4}{1.5cm}{Relational \newline Inference} & CLUTRR & 8 & 1,042 & Q: [Carlos] is [Clarence]'s brother. [Carlos] and his sister, [Annie], went shopping. [Annie] asked her mom [Valerie] if she wanted anything, but [Valerie] said no. How is [Valerie] related to [Clarence]? \newline A: \texttt{``mother''} \\
    \bottomrule    
\end{tabular}
}
\caption{Datasets used for evaluation. ``\# Shot'' stands for the number of few-shot examples in the prompt (following \citet{wei_chain_2022}) and ``\# Test'' stands for the number of test examples.}\label{table:datasets}

 \vspace{-0.1in}
\end{table*}

\subsection{URLs and Licenses}

We use the same distribution of datasets following \citet{wei_chain_2022}:

\paragraph{Math Word Problems}
\begin{itemize}
    \item GSM8K \citep{cobbe_training_2021}: \url{https://github.com/openai/grade-school-math}, MIT license: \url{https://github.com/openai/grade-school-math/blob/master/LICENSE}.
    \item SVAMP \citep{patel_are_2021}: \url{https://github.com/arkilpatel/SVAMP}, MIT license: \url{https://github.com/arkilpatel/SVAMP/blob/main/LICENSE}.
    \item MultiArith \citep{roy_solving_2015}, license: CC BY 4.0.
    \item ASDiv \citep{miao_diverse_2020}: \url{https://github.com/chaochun/nlu-asdiv-dataset}.
    \item AQuA \citep{ling_program_2017}: \url{https://github.com/deepmind/AQuA}, license: \url{https://github.com/deepmind/AQuA/blob/master/LICENSE}.
\end{itemize}

\paragraph{Multi-hop QA}
\begin{itemize}
    \item StrategyQA \citep{geva_did_2021}: we use the open-domain setting (question-only set) from \cite{big-bench_collaboration_beyond_2021}: \url{https://github.com/google/BIG-bench/tree/main/bigbench/benchmark_tasks/strategyqa}.
    \item Date Understanding and Sports Understanding from BIG-Bench \citep{big-bench_collaboration_beyond_2021}: Apache License v.2: \url{https://github.com/google/BIG-bench/blob/main/LICENSE}.
\end{itemize}

\paragraph{Planning}
\begin{itemize}
    \item SayCan \citep{ahn_as_2022}: SayCan dataset can be accessed at \url{https://say-can.github.io/} under CC BY 4.0 license.
\end{itemize}

\paragraph{Relational Reasoning}
\begin{itemize}
    \item CLUTRR \citep{sinha_clutrr_2019}: \url{https://github.com/facebookresearch/clutrr}, license: \url{https://github.com/facebookresearch/clutrr/blob/main/LICENSE}. We obtain the publicly distributed version available at \url{https://drive.google.com/file/d/1SEq_e1IVCDDzsBIBhoUQ5pOVH5kxRoZF/view}, specifically the \texttt{data\_089907f8} split.
\end{itemize}

We use all the above datasets for research purposes only, consistent with their intended use.

\subsection{Data Cleaning}
We perform manual cleaning on ASDiv, Date Understanding, Sports Understanding, and SayCan as we discover a number of annotation issues. In our experiment, we rerun all baselines on our cleaned version of the test sets. They are provided in our repository to assist future research. 

Specifically, we clean each of the datasets as follows: 

\textbf{ASDiv}: We start with the test set used by \citet{wei_chain_2022}, which removes all questions with float-valued and string-valued answers. However, in their released version, we notice an error in the answer extraction step for questions with more than one value in the answer (e.g., ``what is the width and length of X?'', where the answer consists of two values). In their implementation, only the first value is extracted as the ground truth answer, which is then compared against model outputs. This might artificially inflate the final accuracy. To fix this, we extract all values in the answer as a set and compare model outputs against it.

\textbf{Date Understanding}: We find a number of wrong answers in the test set. For example, for the question ``Jane and John married on Jan 2, 1958. It is their 5-year anniversary today. What is the date today in MM/DD/YYYY?'', the provided answer is ``01/02/1961'', whereas the correct answer should be ``01/02/1963''. We manually correct these answers, and the resulting test set has the same number of examples as the original one.

\textbf{Sports Understanding}: We notice a few ambiguities with the Sports Understanding dataset. For instance, running out of bounds is illegal in many sports. The phrase "Domantas Sabonis ran out of bounds" is labeled as implausible, however, Domantas Sabonis is a basketball player, and basketball players can indeed run out of bounds on the court. We remove 8 questions with such action-based ambiguities. Additionally, since the release of this dataset, a few new athletes have risen to fame with identical names to those mentioned in the dataset. For example, the question "Chris Paul struck out the side" is implausible, as the referenced ``Chris Paul'' is a famous basketball player. However, ``Chris Paul'' is also the name of a new MLB baseball player, in which case this statement is plausible. We remove 5 questions with such name-based ambiguities. 

\textbf{SayCan}: We discover a few issues in the test set: (1) the environment setup (e.g., the list of objects, the list of locations, and the initial location of each object) is not the same for all examples; (2) the annotation of the ground truth answer is often incomplete (i.e., for a given task like ``visit all locations'', there exist many possible plans in terms of the order of locations visited, but not all of them are included in the annotation); (3) there are ambiguous descriptions in certain queries, for example, in ``Could you get me something refreshing?'', it is unclear what drinks are considered ``refreshing''. For these questions, we complete the annotation whenever possible, and filter out the rest. The resulting test set contains 103 examples out of the original 120.

\subsection{Dataset Splits}
As stated in Section~\ref{sec:datasets}, we use the official splits whenever possible: training set for exemplar selection, validation set for prompt tuning, and test set for evaluation. In cases where they are available, we adopt the following strategies for each dataset:

\textbf{GSM8K}: it only has training and test sets. We form the validation set by randomly sampling 1,000 examples from the training set.

\textbf{Other MWP datasets}: for AQuA, we use the official training/validation/test split. For the other datasets, only the test sets are used, since we have the same prompt for GSM8K and them.

\textbf{Date Understanding} and \textbf{Sports Understanding}: they only have test sets. We follow \citet{wei_chain_2022} to select the same number of examples from the test set to form the few-shot prompt and use the remaining examples as a new test set. 

\textbf{SayCan}: Following \citet{wei_chain_2022}, we manually write 7 few-shot exemplars, since no training set is provided. We evaluate the models on our cleaned version of the test set, described in the previous subsection.

\textbf{CLUTRR}: this dataset is split into multiple folds. There is a training fold with $K \in \{2,3\}$ (where $K$ is the number of intermediate steps required to reach the answer), and one test fold for each $K$ from 2 to 10. We construct the few-shot prompt using exemplars from the training fold, and test our method on the concatenation of all test folds.

\section{Error Analysis}
\label{sec:appendix_error_analysis}
To further investigate where our method still fails, we inspect 100 errors\footnote{To encourage sample diversity, we embed all the errors using \texttt{text-davinci-002} and cluster the embeddings using spectral clustering. This produces around 70 clusters of different sizes, from which we gather 100 samples using importance sampling.} from model predictions on each of the four datasets and manually annotate the error categories.

\subsection{GSM8K}
\label{sec:appendix_gsm8k_error_analysis}

\begin{figure}[h!]
\centering
\includegraphics[width=0.87\columnwidth]{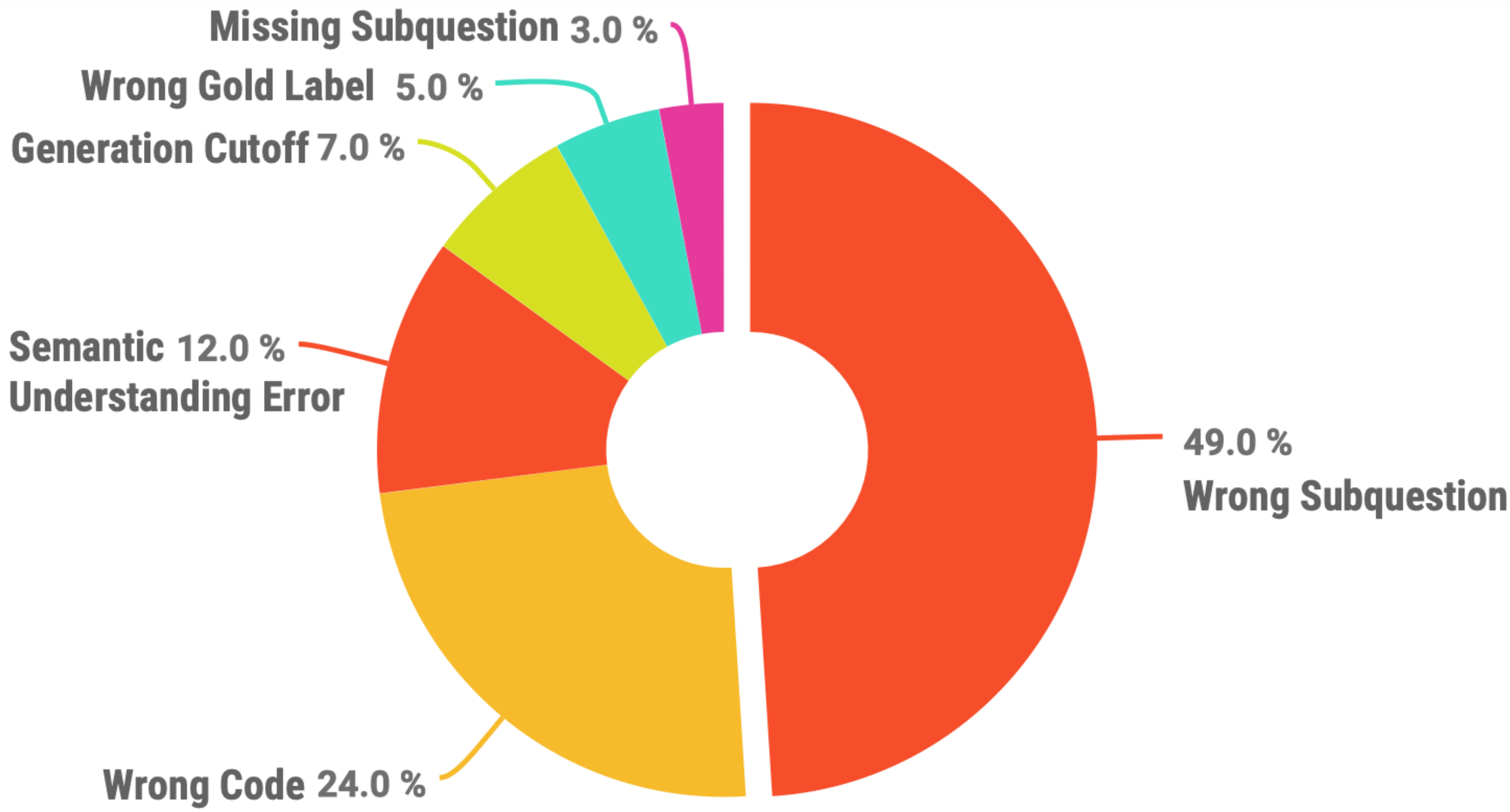}
\vspace{-0.05in}
\caption{Error analysis for GSM8K. For a detailed description of the error categories, see Appendix~\ref{sec:appendix_gsm8k_error_analysis}.}
\vspace{-0.1in}
\label{fig:GSM8K_errors}
\end{figure}

\begingroup
\begin{table*}[h]
    \centering
    \small
    \begin{tabular}{p{0.9\linewidth}}
        \toprule
        \vspace{-2mm}
                  \# 1. How many pounds will Martin lose per week if he eats Cheerios every day for breakfast? (independent, support: ["he'll lose 1.25 pounds/week"])\\
pounds\_lost\_cheerios = 1.25\\
\# 2. How many pounds will Martin gain per week if he eats donuts every day for breakfast? (independent, support: ["he'll gain 1.75 pounds/week"])
pounds\_gained\_donuts = 1.75\\
\# 3. How many weeks are there in 5 weeks? (independent, support: ["External knowledge: there are 7 days in a week"])\\
weeks\_in\_5\_weeks = 5\\
\# 4. How many pounds will Martin lose in 5 weeks if he eats Cheerios every day for breakfast? (depends on 1 and 3, support: [])\\
pounds\_lost\_cheerios\_5\_weeks = pounds\_lost\_cheerios * weeks\_in\_5\_weeks\\
\# 5. How many pounds will Martin gain in 5 weeks if he eats donuts every day for breakfast? (depends on 2 and 3, support: [])\\
pounds\_gained\_donuts\_5\_weeks = pounds\_gained\_donuts * weeks\_in\_5\_weeks\\
\# 6. What will be the difference in his weight at the end of 5 weeks between the two breakfast options? (depends on 4 and 5, support: [])\\
difference\_5\_weeks = pounds\_gained\_donuts\_5\_weeks - pounds\_lost\_cheerios\_5\_weeks\\
\# 7. Final Answer: What will be the difference in his weight at the end of 5 weeks between the two breakfast options? (depends on 6, support: [])\\
answer = difference\_5\_weeks\\
        \bottomrule
    \end{tabular}
       \caption{
    Generated code for the question in Appendix~\ref{sec:appendix_gsm8k_error_analysis}, as an example of "semantic understanding error".
    } 
    \label{tab:appendix-ea}
    
\end{table*}
\endgroup

As shown in Figure~\ref{fig:GSM8K_errors}, we categorize the errors on GSM8K into 6 types, inversely sorted with frequency:\\
\textbf{Wrong Subquestion} (49\%): The LM produces a wrong NL subquestion, which eventually leads to the incorrect answer. While this is the majority error type in our sample, it is worth noting that in a typical human-in-the-loop collaboration, these errors are easily fixable. Even if the user is unfamiliar with programming, they can inspect the NL subquestions and potentially correct the model error by simply deleting or editing a wrong subquestion. \\
\textbf{Wrong Code} (24\%): The NL subquestion is correct, but the code fails to answer the subquestion correctly. For example, the code uses a variable that has not been previously defined.\\
\textbf{Semantic Understanding Error} (12\%): The LM incorrectly interprets certain semantic subtleties in the query. This is the most complex and most interesting error category.
 For example, consider the following problem:
\begin{quote}
    \textit{If Martin eats Cheerios every day for breakfast, he'll \textbf{lose} 1.25 pounds/week. If he eats donuts every day for breakfast, he'll \textbf{gain} 1.75 pounds/week. What will be the difference in his weight at the end of 5 weeks between the two breakfast options?}
\end{quote}
The generated code, as shown in Table~\ref{tab:appendix-ea}, does not assign opposite polarities (signs) for ``pounds lost'' vs. ``pounds gained''. For other examples in this category, we notice errors like missing that a pair of something has 2 items in it, missing to subtract 2 for ``two years ago'' when it occurs as a subjunctive, and so on. Fixing these errors, in general, will require more than providing additional examples in the prompt.\\
\textbf{Generation Cutoff} (7\%): The generation stops midway, mainly due to the LM producing the same steps over and over again. These errors could be easily detected in postprocessing and possibly fixed by re-prompting the LM.\\
\textbf{Wrong Gold Label} (5\%): We find 5 (out of our 100) examples that are genuine annotation errors in the gold labels.\\ 
\textbf{Missing Subquestion} (3\%): The LM misses a relevant subquestion needed for the rest of the reasoning chain to work. These errors are also potentially fixable via human-in-the-loop interaction, where the user can insert a subquestion into the reasoning chain.

\subsection{StrategyQA}
\label{sec:appendix_strategyQA_error_analysis}

\begin{figure}[h!]
\centering
\includegraphics[width=\columnwidth]{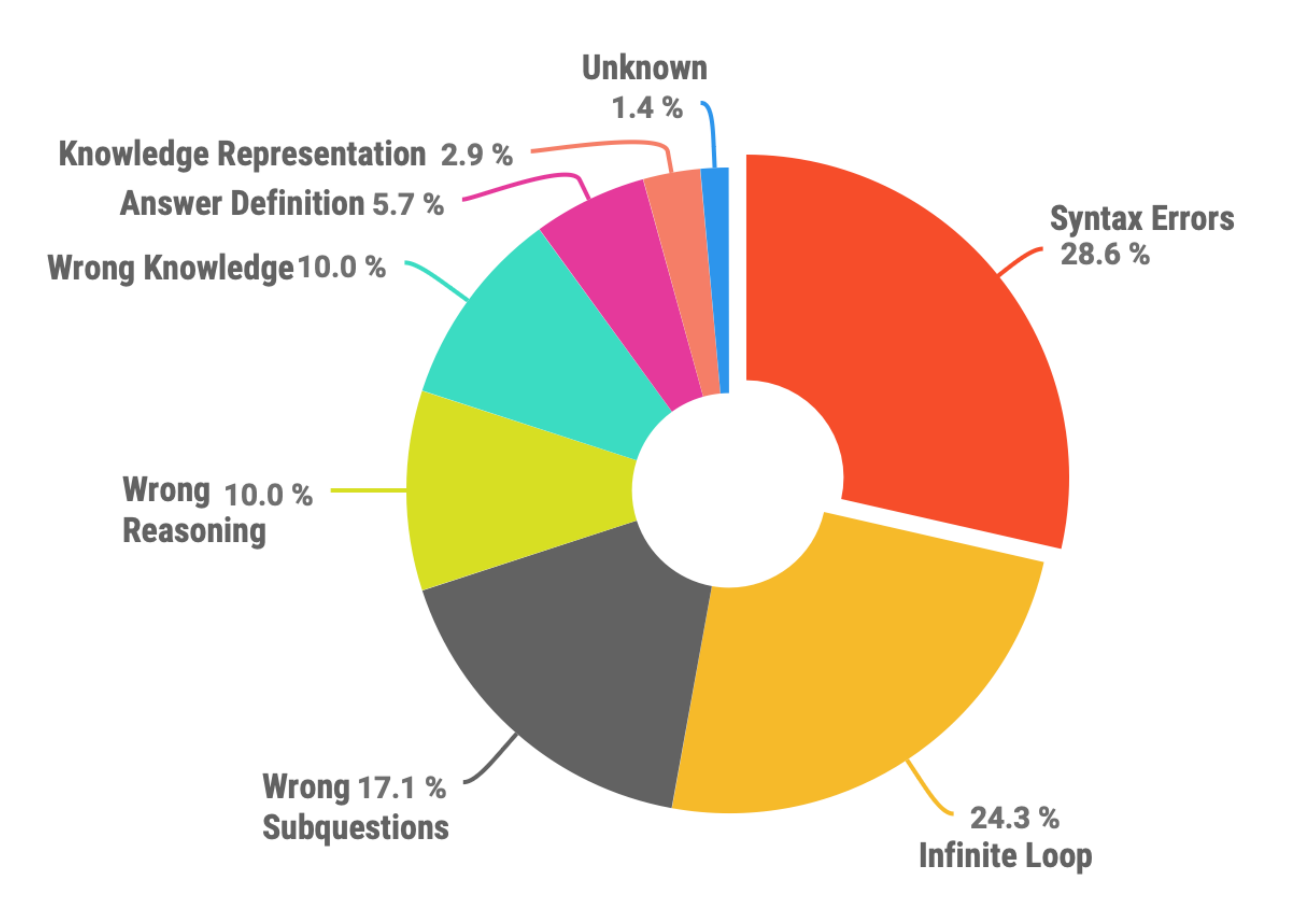}
\vspace{-0.1in}
\caption{Error analysis for StrategyQA. For a detailed description of the error categories, see Section~\ref{sec:appendix_strategyQA_error_analysis}.}
\label{fig:strategyQA_errors}
\end{figure}

As mentioned in Section~\ref{sec:results}, Figure~\ref{fig:strategyQA_errors} shows the error type distribution on a sample of 70 instances from StrategyQA, where we specifically compare the cases where the prediction of CoT is correct whereas ours is wrong. 

There are 8 error types, from most to least frequent:

\textbf{Syntax} (29\%): these are syntax-related errors that result in an invalid Datalog program. For example, the LM redefines a relation that is previously declared, uses an invalid string comparison operation, generates an ungrounded variable in the answer definition, and so on. With further pretraining on Datalog, we believe Codex is able to overcome these errors to a reasonable extent.

\textbf{Infinite Loop} (24\%): the LM generates an infinite loop of subquestions, which are oftentimes rather philosophical. For example, for the question ``Are potatoes native to the European continent?'', the LM generates a list of 40 subquestions (potentially more, if not truncated by the token limit):

\begin{quote}
1. Where are potatoes native to?\\
Potatoes are native to the Andes.\\
2. Which continent is the Andes in?\\
The Andes is in South America.\\
3. Which continent is South America in?\\
South America is in the Americas.\\
4. Which continent is the Americas in?\\
The Americas is in the Western Hemisphere.\\
5. Which hemisphere is the Western Hemisphere in?\\
The Western Hemisphere is in the Earth.\\
6. Which continent is the Earth in?\\
The Earth is in the Solar System.\\
......\\
11. Which continent is the Universe in?\\
The Universe is in the Multiverse.\\
12. Which continent is the Multiverse in?\\
The Multiverse is in the Omniverse.\\
13. Which continent is the Omniverse in?\\
The Omniverse is in the Metaverse.\\
14. Which continent is the Metaverse in?\\
The Metaverse is in the Hyperverse.\\
......\\
39. Which continent is the Quattuordecaverse in?\\
The Quattuordecaverse is in the Quindecaverse.\\
40. Which continent is the Quindecaverse in?
\end{quote}

We notice that this type of errors also happen if the prompt does not include Datalog code but just the NL subquestions, but much less often.

\textbf{Wrong Subquestions} (17\%): The LM generates an irrelevant subquestion to the original question. For instance, given the question, ``Would toast for a vegan have margarine instead of butter?'', one subquestion that is generated is ``What is the difference between a vegan and a vegetarian?'', which does not help with reaching the final answer.

\textbf{Wrong Reasoning} (10\%): The subquestions are correct, but the reasoning from the retrieved facts to the final answer is problematic. For example, to answer the question, ``Do solo pianists require a conductor?'', the model correctly generates the relevant subquestions and their respective answers: 

\begin{quote}
1. What is the definition of a solo pianist?\\
A solo pianist is a pianist who performs alone.\\
2. What is the definition of a conductor?\\
A conductor is a person who leads an orchestra.\\
3. What is the definition of an orchestra?\\
An orchestra is a large instrumental ensemble that contains string, brass, woodwind, and percussion instruments.
\end{quote}

But in deriving the final answer, the reasoning is wrong: ``Now, we derive the final answer: Do solo pianists require a conductor? The answer is Yes only if a solo pianist is not a conductor.''

\textbf{Wrong Knowledge} (10\%): the LM fails to retrieve the correct knowledge to answer the subquestions. For example, given the original question ``Is the largest city in New Mexico also known as Yootó?'', the model correctly generates the subquestions ``What is the largest city in New Mexico?'' (answer: Albuquerque) and ``Is Albuquerque also known as Yootó?''. However, when answering the second subquestion, it retrieves a wrong piece of knowledge (``Albuquerque is also known as Yootó.'', whereas in reality, it should be ``Santa Fe'' that is known as Yootó). 

\textbf{Answer Definition} (6\%): In our prompt, we always derive the answer in the format of ``The answer is Yes only if ...'', which is followed by a Datalog rule containing conditions that should be satisfied for the answer to be true. However, the LM sometimes generates this as ``The answer is No only if ...'', which outputs the reversed answer.

\textbf{Knowledge Representation} (3\%): The retrieved knowledge in NL is correct, but the representation of it in Datalog is wrong. For example, for the piece of knowledge ``The Lucy Show is not the same TV series as JAG (TV series)'', the model represents it as follows:

\begin{small}
    \begin{quote}
    \texttt{.decl Same\_TV\_series(TV\_series1:symbol, TV\_series2:symbol)}\\
    \texttt{Same\_TV\_series("The Lucy Show", "JAG (TV series)").''}
    \end{quote}
\end{small}

which actually means the reverse (they are the same).

\textbf{Unknown} (1\%): There is a very small proportion of errors (1 out of 70) where we are unsure of the cause. Specifically, we expect the solver to output True, but it outputs False instead.

\subsection{Date Understanding}
\label{sec:appendix_date_error_analysis}

\begin{figure}[h]
\centering
\includegraphics[width=\columnwidth]{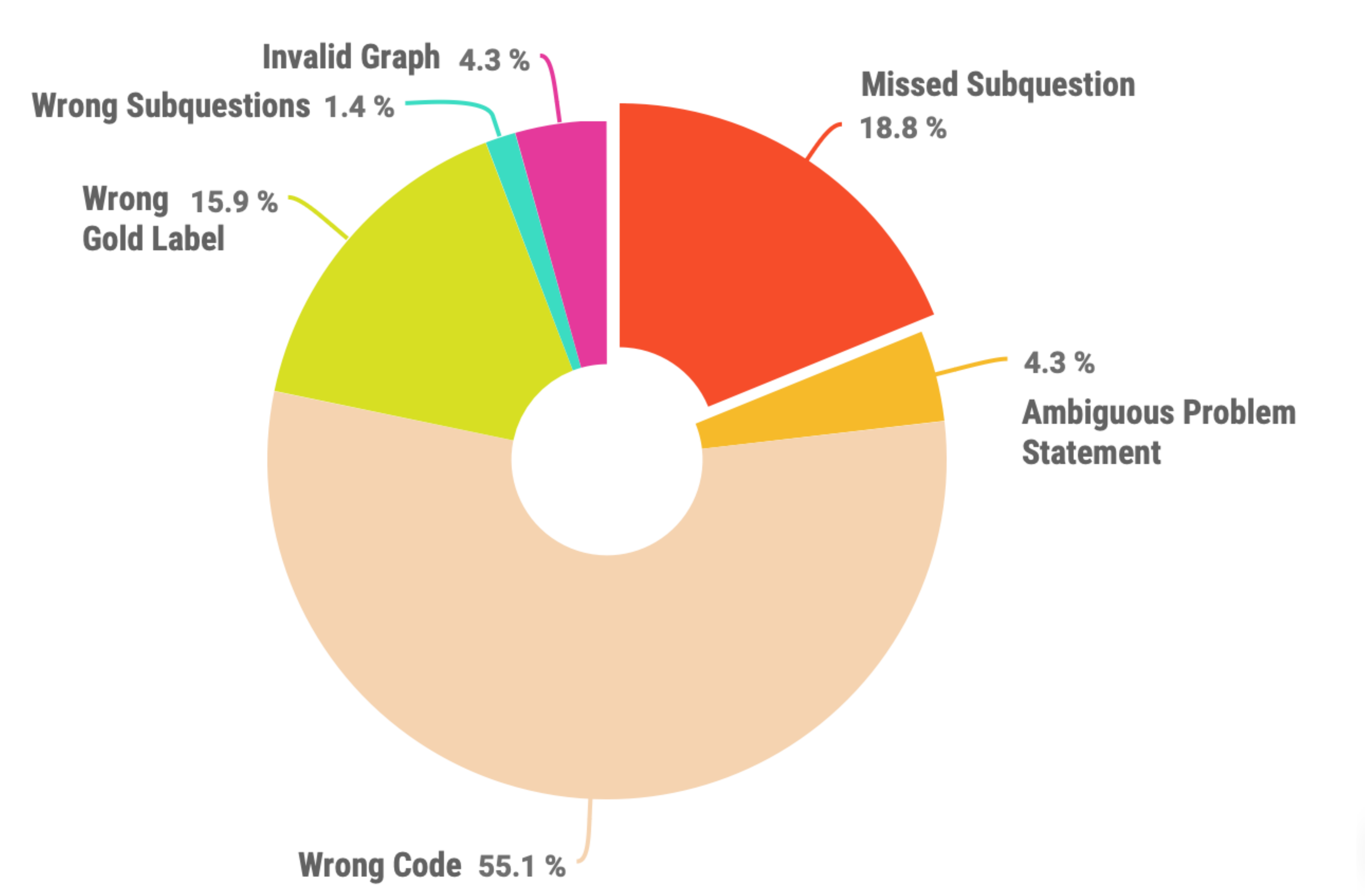}
\vspace{-0.1in}
\caption{Error analysis for Date Understanding. For a detailed description of the error categories, see Appendix~\ref{sec:appendix_date_error_analysis}.}
\vspace{-0.16in}
\label{fig:date_errors}
\end{figure}

Unlike GSM8K, we only have 69 errors out of the 359 test examples, so we annotate them all, as shown in Figure~\ref{fig:date_errors}. The error categories for date understanding are similar to GSM8K, except that we do not see any generation errors in the samples, but we see questions with ambiguous phrasing allowing both the gold and predicted answers to be correct based on interpretation.

\subsection{SayCan}
\label{sec:appendix_saycan_error_analysis}

\begin{figure}[h]
\centering
\includegraphics[width=\columnwidth]{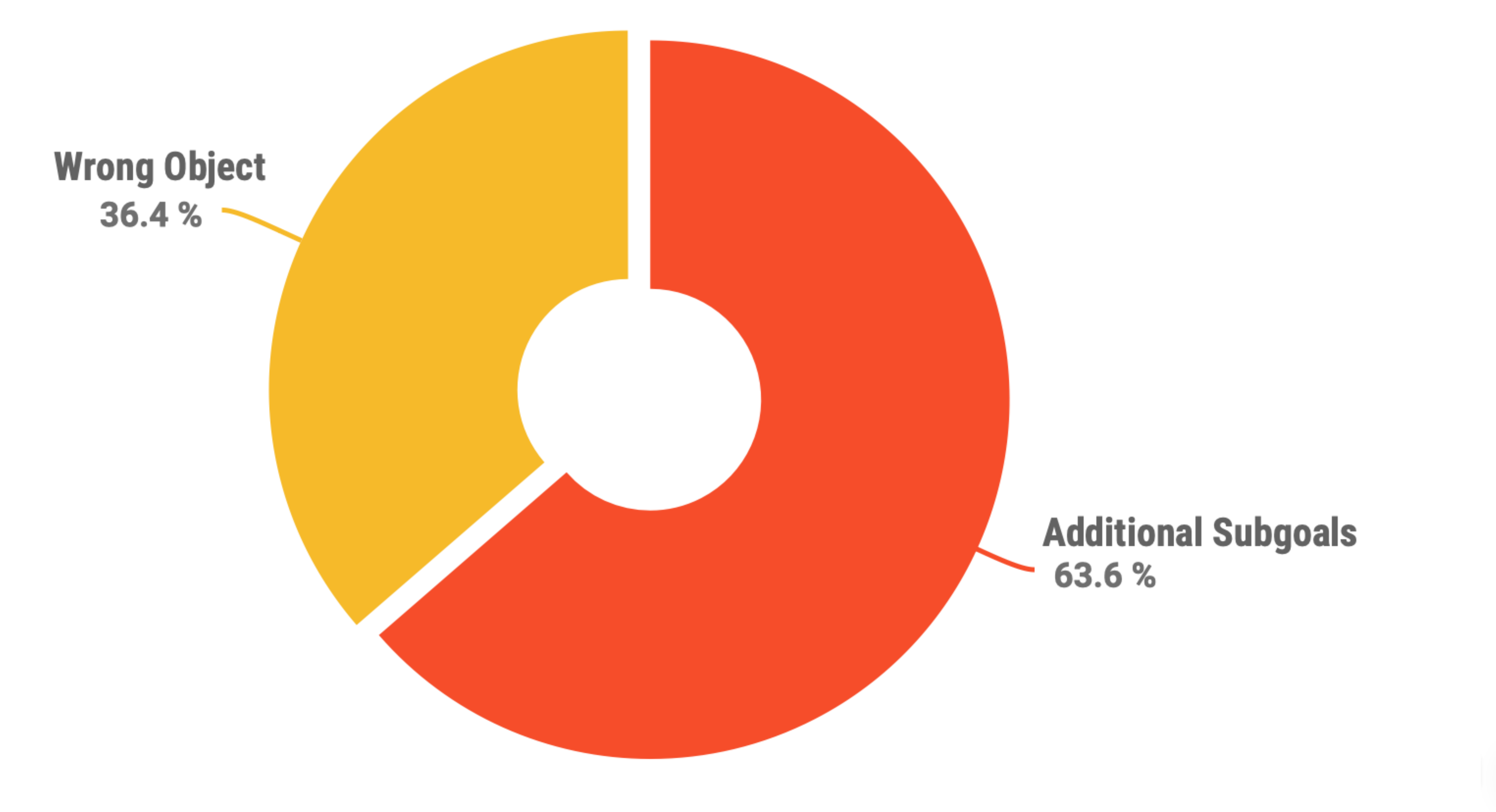}
\vspace{-0.1in}
\caption{Error analysis for SayCan. For a detailed description of the error categories, see Section~\ref{sec:appendix_saycan_error_analysis}.}
\vspace{-0.16in}
\label{fig:saycan_errors}
\end{figure}

Since SayCan only has 120 test examples and Faithful CoT produces 7 errors, we annotate all 7 of them, as shown in Figure~\ref{fig:saycan_errors}. These 7 examples can be categorized into two types:\\
\textbf{Additional Subgoals} (64\%): Cases where the model generated unnecessary subgoals in the decomposition of the original task, leading the planner astray. This is illustrated by the request ``Clear the jalapeno chips off the counter'':

\begin{small}
\begin{verbatim}
    (:goal
        (and
                (not (at jalapeno-chips counter))
                (not (at jalapeno-chips table))
                (not (at jalapeno-chips trash))
                (not (at jalapeno-chips bowl))
                (not (at jalapeno-chips user))
        )
    )
\end{verbatim}
\end{small}

\textbf{Wrong Object} (36\%): Here the model generates the wrong object/object types in the goal. For example, a request such as ``I opened a pepsi earlier. How would you bring me an open can?'' fails because the model generates actions with water instead of Pepsi.

\subsection{CLUTRR}
\label{sec:appendix_clutrr_error_analysis}

\begin{figure}[h]
\centering
\includegraphics[width=\columnwidth]{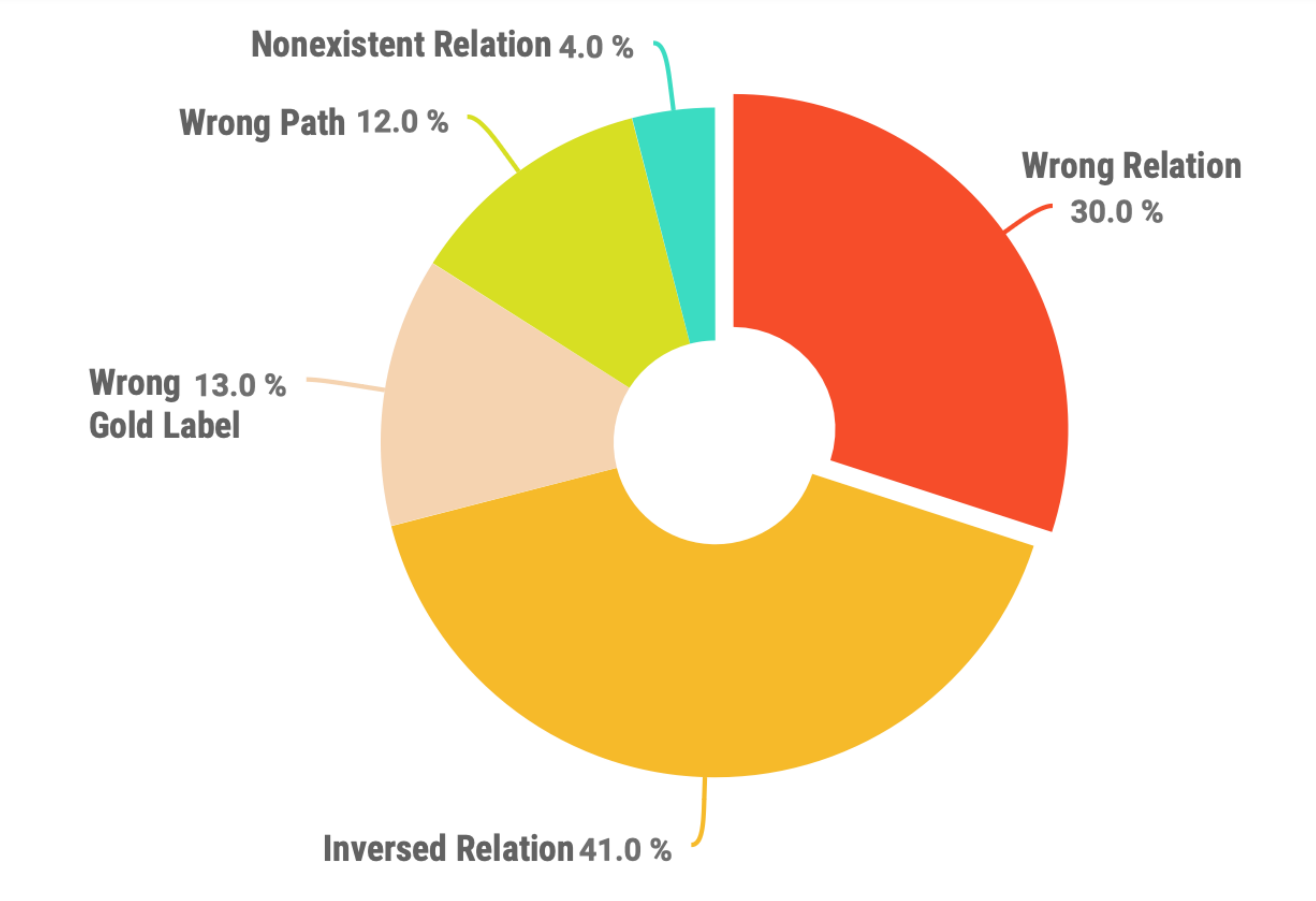}
\vspace{-0.1in}
\caption{Error analysis for CLUTRR. For a detailed description of the error categories, see Section~\ref{sec:appendix_clutrr_error_analysis}.}
\vspace{-0.16in}
\label{fig:clutrr_errors}
\end{figure}

For CLUTRR, we group all error cases by $K$, the number of steps in their gold reasoning chain, as a proxy for problem complexity, and perform importance sampling on these groups to select 100 examples. Our annotation of these examples reveals 5 error categories, as shown in Figure~\ref{fig:clutrr_errors}:\\
\textbf{Inversed Relation} (41\%): This stands out as the majority of the errors. These errors are caused by the reversal of directional relationships for the actors in the problem, i.e., predicting ``mother'' or ``nephew'' when the answer is ``daughter'' or ``uncle'' respectively.\\
\textbf{Wrong Relation} (30\%): Here the model extracts the relation incorrectly (not even the inverse). For example, for the subquestion ``How is [Donald] related to [Jason]?'' with the correctly identified support ``[Jason] is father of their father'', the model produces \texttt{relation(Donald, Jason) = son} when the correct relation should be ``grandson''.\\
\textbf{Nonexistent Relation} (4\%): The model hallucinates a non-existent relation (e.g. ``adopted'' for daughter). \\
\textbf{Wrong Path} (12\%): Here, the model does not generate a correct reasoning path from target entity A to target entity B in the question.\\
\textbf{Wrong Gold Label} (13\%): These are annotation errors in the CLUTRR dataset. In one example, for the sentence, ``[Gloria] asked her mother [Laura] if she could go outside and play with her friends.'', the annotation says Laura is Gloria's grandmother.

\section{Prompts}
\label{sec:appendix_prompt}

Due to the space limit, we show one exemplar in the prompt for each dataset here. The full prompts can be found in our repository.

Among all the MWP datasets, our prompt for AQuA is different from the rest, because the answers are in a multiple-choice format instead of integers. To produce a multiple-choice answer, we take a two-step approach by first producing a numerical answer in the same way as for the other math datasets. Then, we perform an additional step of converting the numerical answer into an answer choice by again prompting the language model to generate which answer choice is closest to the previously produced numerical answer. An exemplar of this 2-step prompt is shown in Table~\ref{tab:appendix-AQUA_prompt}.

\begingroup
\begin{table*}[!t]
    \centering
    \small
    \scalebox{0.9}{
    \begin{tabular}{p{\linewidth}}
        \toprule
        \underline{\textbf{\textsc{Exemplar for AQuA}}} \\
        \vspace{-1mm}
        \underline{Step 1: Answer Prediction}\\
        \# Question: In a flight of 600 km, an aircraft was slowed down due to bad weather. Its average speed for the trip was reduced by 200 km/hr and the time of flight increased by 30 minutes. The duration of the flight is: \\
        \vspace{-1mm}
        \# Answer option: ['A)1 hour', 'B)2 hours', 'C)3 hours', 'D)4 hours', 'E)5 hours'] \\
        \# Write Python code to solve the following questions. Store your result as a variable named 'answer'. \\
        \# 1. What was the duration of the flight? (independent, support: ["The duration of the flight is"]) \\
        duration = Symbol('duration', positive=True) \\
        \# 2. What is the delay of the flight? (independent, support: ["the time of flight increased by 30 minutes"]) \\
        delay = 30 / 60 \\
        \# 3. What was the total flight distance? (independent, support: ["In a flight of 600 km"]) \\
        total\_distance = 600 \\
        \# 4. What was the original speed? (depends on 1 and 3, support: ["External knowledge: speed is distance over time"]) \\
        original\_speed = total\_distance / duration \\
        \# 5. What was the reduced speed? (depends on 1, 2, and 3, support: []) \\
        reduced\_speed = total\_distance / (duration + delay) \\
        \# 6. What was the duration of the flight if the original speed was 200 km/hr faster than the reduced speed? (depends on 4, 5, and 1, support: []) \\
        solution = solve\_it(original\_speed - reduced\_speed - 200, duration) \\
        answer = solution[duration] \\
        
        \vspace{2mm}

        \underline{Step 2: Multiple Choice Conversion}\\
        \# Question: In a flight of 600 km, an aircraft was slowed down due to bad weather. Its average speed for the trip was reduced by 200 km/hr and the time of flight increased by 30 minutes. The duration of the flight is: \\
        \vspace{-1mm}
        \# Answer option: ['A)1 hour', 'B)2 hours', 'C)3 hours', 'D)4 hours', 'E)5 hours'] \\
        \# Prediction: 1.00000000000000 \\
        \# Closest Option: A \\
        \bottomrule
    \end{tabular}
    }
        \caption{
    An exemplar from our prompt for AQuA.
    }
    \label{tab:appendix-AQUA_prompt}
\end{table*}
\endgroup
\begingroup
\begin{table*}[!t]
    \centering
    \small
    \scalebox{0.9}{
    \begin{tabular}{p{\linewidth}}
        \toprule
        \underline{\textbf{\textsc{Exemplar for GSM8K, SVAMP, MultiArith, and ASDiv}}} \\
        \vspace{-2mm}
        \# Q: There are 15 trees in the grove. Grove workers will plant trees in the grove today. After they are done, there will be 21 trees. How many trees did the grove workers plant today? \\
        \vspace{-1mm}
        \# To answer this question, write a Python program to answer the following subquestions: \\
        \# 1. How many trees are there in the beginning? (independent, support: ["There are 15 trees"]) \\
        trees\_begin = 15 \\
        \# 2. How many trees are there in the end? (independent, support: ["there will be 21 trees"]) \\
        trees\_end = 21 \\
        \# 3. How many trees did the grove workers plant today? (depends on 1 and 2, support: []) \\
        trees\_today = trees\_end - trees\_begin \\
        \# 4. Final Answer: How many trees did the grove workers plant today? (depends on 3, support: []) \\
        answer = trees\_today \\
        \bottomrule
    \end{tabular}
    \label{tab:appendix-MWP_prompt}
    }
        \caption{
    An exemplar from our prompt for GSM8K, SVAMP, MultiArith, and ASDiv.
    }
\end{table*}
\endgroup
\begingroup
\begin{table*}[!t]
    \centering
    \small
    \scalebox{0.9}{
    \begin{tabular}{p{\linewidth}}
        \toprule
        \underline{\textbf{\textsc{Exemplar for StrategyQA}}} \\
        \vspace{-2mm}
        // Q: Would a pear sink in water? \\
        \vspace{-1mm}
        // To answer this question, we answer the following subquestions: \\
        // 1. What is the density of a pear? \\
        // The density of a pear is about $0.6 g/cm^3$. \\
        // 2. What is the density of water? \\
        // Water has a density of $1 g/cm^3$. \\ \\
        
        // Then, we represent these answers in Datalog: \\
        // 1. The density of a pear is about $0.6 g/cm^3$. \\
        .decl Has\_density(Object:symbol, Density:float) \\
        Has\_density("pear", 0.6). \\
        // 2. Water has a density of $1 g/cm^3$. \\
        Has\_density("water", 1). \\ \\
        
        // Now, we derive the final answer: Would a pear sink in water? \\
        // The answer is Yes only if the density of a pear is more than the density of water. \\
        .decl Answer() \\
        Answer() :- Has\_density("pear", density1), Has\_density("water", density2), density1 > density2. \\
        .output Answer \\
        \bottomrule
    \end{tabular}
    }
        \caption{
    An exemplar from our prompt for StrategyQA.
    }
    \label{tab:appendix-StrategyQA_prompt}
\end{table*}
\endgroup
\begingroup
\begin{table*}[!t]
    \centering
    \small
    \scalebox{0.9}{
    \begin{tabular}{p{\linewidth}}
        \toprule
        \underline{\textbf{\textsc{Exemplar for Date Understanding}}} \\
        \vspace{-2mm}
        \# Q: Yesterday was April 30, 2021. What is the date tomorrow in MM/DD/YYYY? \\
        \vspace{-1mm}
        \# To answer this question, we write a program to answer the following subquestions: \\
        \# import relevant packages \\
        from datetime import date, time, datetime \\
        from dateutil.relativedelta import relativedelta \\
        \# 1. What is the date yesterday? (independent, support: ["Yesterday was April 30, 2021"]) \\
        date\_yesterday = date(2021,4,30) \\
        \# 2. What is the date today? (depends on 1, support: ["Yesterday was April 30, 2021"]) \\
        date\_today = date\_yesterday + relativedelta(days=1) \\
        \# 3. What is the date tomorrow? (depends on 2, support: []) \\
        date\_tomorrow = date\_today + relativedelta(days=1) \\
        \# 4. Final Answer: What is the date tomorrow in MM/DD/YYYY? (depends on 3, support: []) \\
        answer = date\_tomorrow.strftime("\%m/\%d/\%Y") \\
        \bottomrule
    \end{tabular}
    }
        \caption{
    An exemplar from our prompt for Date Understanding.
    }
        \label{tab:appendix-date_prompt}

\end{table*}

\endgroup
\begingroup
\begin{table*}[!t]
    \centering
    \small
    \scalebox{0.9}{
    \begin{tabular}{p{\linewidth}}
        \toprule
        \underline{\textbf{\textsc{Exemplar for Sports Understanding}}} \\
        \vspace{-2mm}
        \# Q: Is the following statement plausible? Sam Darnold passed the puck \\
        \vspace{-1mm}
        \# To answer this question, write a Python program to answer the following subquestions: \\
        \# 1. Sam Darnold is a player in which sport? (independent, support: ["Sam Darnold is an  NFL Quarterback", "NFL is the National Football League"]) \\
        player\_sport = "football" \\
        \# 2. The phrase "passed the puck" implies playing which sport? (independent, support: ["Players pass the puck in hockey"]) \\
        playing\_sport = "hockey" \\
        \# 3. Is the following statement plausible? Sam Darnold passed the puck (depends on 1 and 2, support: ["Sam Darnold is an  NFL Quarterback", "NFL is the National Football League", "Players pass the puck in hockey"]) \\
        plausibility = (player\_sport == playing\_sport) \\
        \# 4. Is the following statement plausible? Sam Darnold passed the puck (depends on 3, support: []) \\
        answer = int(plausibility) \\
        \bottomrule
    \end{tabular}
    }
        \caption{
    An exemplar from our prompt for Sports Understanding.
    }
    \label{tab:appendix-sports_prompt}
\end{table*}
\endgroup
\begingroup
\begin{table*}[!t]
    \centering
    \small
    \scalebox{0.9}{

    \begin{tabular}{p{\linewidth}}
        \toprule
        \underline{\textbf{\textsc{Exemplar for SayCan}}} \\
        \vspace{-2mm}
        User query: Bring me something not sweet to eat. \\
        \vspace{-1mm}
        Goal in PDDL: \\
        (:goal \\
        	\hspace{0.2in} ; I need to find a snack \\
        	\hspace{0.2in} (exists (?s - snack) \\
        		\hspace{0.2in} \hspace{0.2in} ; it has to satisfy the following conditions \\
        		\hspace{0.2in} \hspace{0.2in} (and \\
        			\hspace{0.2in} \hspace{0.2in} \hspace{0.2in} ; the snack must not be sweet \\
        			\hspace{0.2in} \hspace{0.2in} \hspace{0.2in} (not (is-sweet ?s)) \\
        			\hspace{0.2in} \hspace{0.2in} \hspace{0.2in} ; bring it to the user \\
        			\hspace{0.2in} \hspace{0.2in} \hspace{0.2in} (at ?s user) \\
        		\hspace{0.2in} \hspace{0.2in} ) \\
        	\hspace{0.2in} ) \\
        ) \\
        \bottomrule
    \end{tabular}
    }
    \label{tab:appendix-saycan_prompt}
        \caption{
    An exemplar from our prompt for SayCan.
    }
\end{table*}
\endgroup
\begingroup
\begin{table*}[!t]
    \centering
    \small
    \scalebox{0.9}{
    \begin{tabular}{p{\linewidth}}
        \toprule
        \underline{\textbf{\textsc{Exemplar for CLUTRR}}} \\
        \vspace{-2mm}
        \# Context: [Jason] always had some great adventure planned for his granddaughter [Guillermina] when she came to visit. So, naturally, when [Myrna] told her daughter [Guillermina] that they would be going to visit [Jason] she could hardly contain herself. \\
        \vspace{-1mm}
        \# Question: How is [Jason] related to [Myrna]? \\
        \# To answer this question, we write a program to answer the following subquestions: \\
        \# 1. How is [Jason] related to [Guillermina]? (independent, support: "[Jason] always had some great adventure planned for his granddaughter [Guillermina] when she came to visit.") \\
        relation(Jason, Guillermina) = grandfather \\
        \# 2. How is [Guillermina] related to [Myrna]? (independent, support: "So, naturally, when [Myrna] told her daughter [Guillermina] that they would be going to visit [Jason] she could hardly contain herself.") \\
        relation(Guillermina, Myrna) = daughter \\
        \# 3. Final answer: How is [Jason] related to [Myrna]? (depends on 1, 2) \\
        relation(Jason, Myrna) = relation(Jason, Guillermina) @ relation(Guillermina, Myrna) \\
        \bottomrule
    \end{tabular}
    }
        \caption{
    An exemplar from our prompt for CLUTRR.
    }
    \label{tab:appendix-CLUTRR_prompt}
\end{table*}
\endgroup

\end{document}